\documentclass[journal,twocolumn]{IEEEtran}
\usepackage{xcolor,soul}
\usepackage{mathtools,comment}
\usepackage{epsfig,makeidx,color}
\usepackage{amsmath,amssymb,bbm,enumitem}
\usepackage{cite,graphicx,lipsum}
\usepackage[switch,pagewise]{lineno}
\usepackage{tabularx}
\usepackage{algorithm}
\usepackage{algpseudocode}
\usepackage{hyperref}
\usepackage{cases}
\usepackage{graphicx}
\usepackage{booktabs}
\usepackage{multirow}
\usepackage{dsfont}
\usepackage{subcaption}
\usepackage{cleveref}
\usepackage{tablefootnote}
\usepackage{booktabs}
\usepackage{caption}
\usepackage{tikz}

\usepackage{subcaption}

\hypersetup{
        colorlinks = true,
        citecolor = red,
        breaklinks = true,
        linktocpage=true,
}
\usepackage{url}

\def\cM{{\cal M}}

\def\cK{{\cal K}}
\def\cX{{\cal X}}
\def\cY{{\cal Y}}
\def\cG{{\cal G}}

\def\rT{{\rm T}}

\def\uR{{\mathbb R}}

\def\uE{{\mathbb E}}

\def\uN{{\mathbb N}}

\def\be{ \begin{equation} }
\def\ee{ \end{equation} }
\def\bea{ \begin{eqnarray} }
\def\eea{ \end{eqnarray} }

\def\bx{{\bf x}}

\def\bg{{\bf g}}

\def\bu{{\bf u}}

\def\bh{{\bf h}}

\def\bz{{\bf z}}

\def\bw{{\bf w}}

\def\bF{{\bf F}}

\def\bK{{\bf K}}

\def\b0{{\bf 0}}

\def\cD{{\cal D}}

\ifCLASSOPTIONonecolumn
\else

\fi

\newcommand{\mybar}[2]{%
  \makebox[2cm][l]{%
    \begin{tikzpicture}[baseline]
      \pgfmathsetmacro{\value}{#1/0.003*1} 
      \definecolor{barcolor}{rgb}{0.6,0.6,1} 
      \ifnum#2=1\relax
        \definecolor{barcolor}{rgb}{0,1,0} 
        \node[anchor=west, font=\bfseries] at (\value,0.1) {#1}; 
      \fi
      \draw[fill=barcolor, draw=black, opacity=0.8] (0,0) rectangle (\value,0.2); 
    \end{tikzpicture}%
  }%
}

\newcommand{\myredbar}[2]{%
  \makebox[2cm][l]{%
    \begin{tikzpicture}[baseline]
      \pgfmathsetmacro{\value}{#1/0.4*1} 
      \definecolor{barcolor}{rgb}{1,0.6,0.6} 
      \ifnum#2=1\relax
        \definecolor{barcolor}{rgb}{0,1,0} 
        \node[anchor=west, font=\bfseries] at (\value,0.1) {#1}; 
      \fi
      \draw[fill=barcolor, draw=black, opacity=0.8] (0,0) rectangle (\value,0.2); 
    \end{tikzpicture}%
  }%
}

\newcommand{\normalbar}[3]{%
  \makebox[3cm][l]{%
    \begin{tikzpicture}[baseline]
      \pgfmathsetmacro{\mean}{#1/0.3*3} 
      \pgfmathsetmacro{\ci}{#2/0.3*3}
      \definecolor{meancolor}{rgb}{0.7,0.7,0.7} 
      \definecolor{cibarcolor}{rgb}{0.7,0.7,0.7} 
      \ifnum#3=1\relax
        \node[anchor=west, font=\bfseries] at (\mean,0.1) {\textbf{#1 ± #2}}; 
      \fi
      \draw[fill=meancolor, draw=black, opacity=0.8] (0,0) rectangle (\mean,0.2); 
      \draw[fill=cibarcolor, draw=black, opacity=0.8] (\mean-\ci,0.075) rectangle (\mean+\ci,0.125); 
    \end{tikzpicture}%
  }%
}

\newcommand{\mybarThreeD}[2]{%
  \makebox[2cm][l]{%
    \begin{tikzpicture}[baseline]
      \pgfmathsetmacro{\value}{#1/0.003*1} 
      \definecolor{barcolor}{rgb}{0.6,0.6,1} 
      \ifnum#2=1\relax
        \definecolor{barcolor}{rgb}{0,1,0} 
        \node[anchor=west, font=\bfseries] at (\value,0.1) {#1}; 
      \fi
      \draw[fill=barcolor, draw=black, opacity=0.8] (0,0) rectangle (\value,0.2); 
    \end{tikzpicture}%
  }%
}

\newcommand{\myredbarThreeD}[2]{%
  \makebox[2cm][l]{%
    \begin{tikzpicture}[baseline]
      \pgfmathsetmacro{\value}{#1/0.4*1} 
      \definecolor{barcolor}{rgb}{1,0.6,0.6} 
      \ifnum#2=1\relax
        \definecolor{barcolor}{rgb}{0,1,0} 
        \node[anchor=west, font=\bfseries] at (\value,0.1) {#1}; 
      \fi
      \draw[fill=barcolor, draw=black, opacity=0.8] (0,0) rectangle (\value,0.2); 
    \end{tikzpicture}%
  }%
}

\newcommand{\normalbarThreeD}[3]{%
  \makebox[3cm][l]{%
    \begin{tikzpicture}[baseline]
      \pgfmathsetmacro{\mean}{#1/0.3*3} 
      \pgfmathsetmacro{\ci}{#2/0.3*3}
      \definecolor{meancolor}{rgb}{0.7,0.7,0.7} 
      \definecolor{cibarcolor}{rgb}{0.7,0.7,0.7} 
      \ifnum#3=1\relax
        \node[anchor=west, font=\bfseries] at (\mean,0.1) {\textbf{#1 ± #2}}; 
      \fi
      \draw[fill=meancolor, draw=black, opacity=0.8] (0,0) rectangle (\mean,0.2); 
      \draw[fill=cibarcolor, draw=black, opacity=0.8] (\mean-\ci,0.075) rectangle (\mean+\ci,0.125); 
    \end{tikzpicture}%
  }%
}

\newcommand{\sharedxaxismin}[1]{%
  \makebox[2cm][l]{%
    \begin{tikzpicture}[baseline]
      \draw[->] (0,0.2) -- (2,0.2); 
      \foreach \x/\xtext in {0/0, 2/0.003} {
        \draw (\x,0.2) -- (\x,0.1) node[below] {\xtext};
      }
    \end{tikzpicture}%
  }%
}

\newcommand{\sharedxaxismax}[1]{%
  \makebox[2cm][l]{%
    \begin{tikzpicture}[baseline]
      \draw[->] (0,0.2) -- (2,0.2); 
      \foreach \x/\xtext in {0/0, 2/0.4} {
        \draw (\x,0.2) -- (\x,0.1) node[below] {\xtext};
      }
    \end{tikzpicture}%
  }%
}

\newcommand{\sharedxaxisconf}[1]{%
  \makebox[3cm][l]{%
    \begin{tikzpicture}[baseline]
      \draw[->] (0,0.2) -- (3,0.2); 
      \foreach \x/\xtext in {0/0, 3/0.3} {
        \draw (\x,0.2) -- (\x,0.1) node[below] {\xtext};
      }
    \end{tikzpicture}%
  }%
}

\begin{document}

\title{Predictive Covert Communication Against Multi-UAV Surveillance Using Graph Koopman Autoencoder

\thanks{$^\dagger$S. Krishnan and J. Choi are with the School of Information Technology, Deakin University, Australia.}
\thanks{$^*$J. Park is with the Information Systems Technology and Design pillar, Singapore University of Technology and Design, Singapore.}
\thanks{$^\ddagger$G. Sherman and B. Campbell are with the Mission Autonomy, Platforms Division, Defence Science and Technology Group, Australia.}

\author{%
Sivaram Krishnan$^\dagger$, Jihong Park$^*$, Gregory Sherman$^\ddagger$, Benjamin Campbell$^\ddagger$, and Jinho Choi$^\dagger$
}}


\maketitle
\begin{abstract}
Low Probability of Detection (LPD) communication aims to obscure the presence of radio frequency (RF) signals to evade surveillance. In the context of mobile surveillance utilizing unmanned aerial vehicles (UAVs), achieving LPD communication presents significant challenges due to the UAVs' rapid and continuous movements, which are characterized by unknown nonlinear dynamics. Therefore, accurately predicting future locations of UAVs is essential for enabling real-time LPD communication. In this paper, we introduce a novel framework termed predictive covert communication, aimed at minimizing detectability in terrestrial ad-hoc networks under multi-UAV surveillance. Our data-driven method synergistically integrates graph neural networks (GNN) with Koopman theory to model the complex interactions within a multi-UAV network and facilitating long-term predictions by linearizing the dynamics, even with limited historical data. Extensive simulation results substantiate that the predicted trajectories using our method result in at least 63\%-75\% lower probability of detection when compared to well-known state-of-the-art baseline approaches, showing promise in enabling low-latency covert operations in practical scenarios. 
\end{abstract}


\ifCLASSOPTIONonecolumn
\baselineskip 28pt
\fi

\section{Introduction}

While wireless communication can provide tremendous benefits, there are several challenges in military applications. In general, it is required that communication be secure \cite{Qian2021}, which can be accomplished using cryptographic methods \cite{Menezes96}. Although traditional cryptographic methods are effective in obscuring the content of messages through various encryption techniques, they fail to address the risk of detecting the communication itself. Thus, it is also required to support covert or low probability of detection (LPD) communication capabilities \cite{bash2015hiding}.
The research on covert communication can be categorized into the following two main areas:
\begin{itemize}
    \item The information-theoretic aspects \cite{bash2015hiding, sobers2017covert}, which focus on identifying the fundamental limits of covert communication. The key research question here is, "\emph{How much information can a transmitter send to a receiver while ensuring covert communication in the presence of an eavesdropper?}"
    \item The second area of research concerns the optimizing of communication parameters such as transmission power \cite{carmi2006minimum}, modulation schemes \cite{hijaz2010exploiting}, bandwidth allocation \cite{mao2023joint}. The optimization problem might consider one or more parameters. This involves questions such as, "\emph{What communication parameters should the transmitter use to ensure that the received signal-to-noise ratio (SNR) is below the noise floor?} \cite{shakeel2023gaussian, krishnan2024graph}".
\end{itemize}

Within the second category, covert communication for wireless ad-hoc networks is addressed with power control in an area of operation (AO) to support LPD communication in \cite{campbell2018minimising} \cite{Krishnan_SSP23}. Additionally, \cite{Sims19} proposes a distributed topology control algorithm for LPD communication in wireless ad-hoc networks.

In covert communication, it is crucial to understand the capabilities of eavesdroppers or wardens who attempt to detect the presence of radio frequency (RF) signals. When considering an AO where nodes in a wireless ad-hoc network wish to communicate using LPD techniques, wardens may exploit their mobility, if possible, rather than remaining in fixed locations. This mobility makes it more challenging to maintain covert communication compared to dealing with static wardens. In this context, unmanned aerial vehicles (UAVs) can be considered for surveillance \cite{Memos21} \cite{Bist21} with RF sensing capabilities.

In this paper \footnote{This paper is an extended version of \cite{krishnan2024graph}, where the initial ideas of the paper were presented at the IEEE 99th Vehicular Technology Conference (VTC 2024 - Spring).}, we holistically address the problem of enabling covert communication, diverging from the traditional scenario characterized by point-to-point (P2P) communication between Alice and Bob, where an eavesdropping warden named Willie seeks to intercept their transmission. Instead, we focus on covert communication for wireless ad-hoc networks under UAV surveillance, extending the scenario from \cite{campbell2018minimising}. In such environments, numerous Alice(s) and Bob(s) strive to communicate while multiple mobile Willie(s) collaborate to cover the AO, detecting any presence of RF signals. Specifically, considering modern surveillance equipment, we extend Willie(s) to a multi-UAV network \cite{semsch2009autonomous}. This network is deployed for efficient surveillance purposes, leveraging the advantages of UAVs, such as their size and detectability. The multi-UAV network is equipped with sensors and communication interception capabilities to monitor various aspects of communication, including the signal strength of transmissions at different locations. 

\begin{figure}
    \centering
\includegraphics[width=1 \columnwidth]{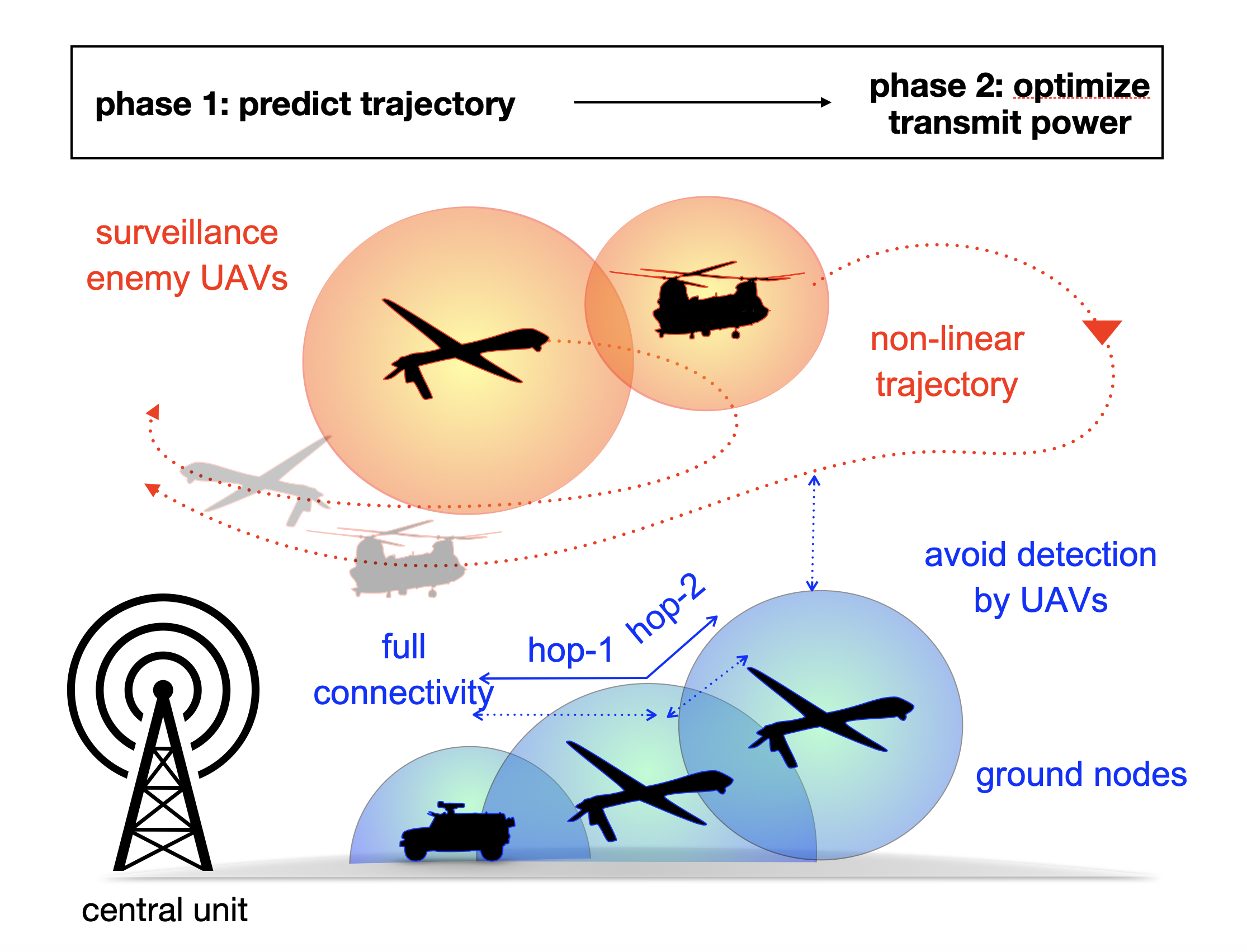}
    \caption{Schematic representation of the proposed predictive covert communication framework, illustrating proactive transmit power assignment to ground nodes by a central unit that predicts long-term trajectories for the multi-UAV network using historical location data.} 
    \label{fig:problem}
\end{figure}

While the multi-UAV network offers significant surveillance advantages, it becomes challenging for wireless ad-hoc networks to maintain covert communication, especially due to the UAVs' mobility. Therefore, an external entity is needed to assist the nodes in the ad-hoc network with covert communication. To this end, a central unit (CU) is considered, where the locations of the UAVs are monitored to aid the nodes of a covert ad-hoc network. The proposed framework is illustrated in Fig.~\ref{fig:problem}, depicting an AO with a wireless ad-hoc network, multi-UAV surveillance, and a CU. Furthermore, we make the following assumptions:
\begin{itemize}
\item[\textbf{A1)}] The CU's location is known, and it has access to the past locations of the UAV network, which have been accurately monitored and collected using radar techniques.
\item[\textbf{A2)}] The CU has sufficient processing capabilities to predict the long-term trajectories of the UAVs and broadcast the transmit power to the nodes for covert operations. While the RF signal of the CU can be detected, it does not affect the covert operations, as the CU is located outside of the AO.
\end{itemize}

While the CU can aid covert communication under UAV surveillance, various challenges exist. The multi-UAV network is \emph{temporal} in nature, meaning its location changes over time and is governed by a set of nonlinear dynamical equations. Therefore, simple prediction techniques based on linear models may not provide reasonable performance. In addition, the nonlinear dynamics of multiple UAVs are generally unknown, making model-based prediction difficult. Thus, in this paper, we propose using a data-driven technique for predicting multi-UAV trajectories, based on recent developments in data-driven techniques applied to complex systems \cite{Brunton_Kutz_2022}. The contributionsof the paper are summarized as follows:
\begin{itemize}
\item[\textbf{C1)}] We introduce a novel framework for enhancing covert wireless ad-hoc network communications under the surveillance of dynamically moving UAVs. Our approach, termed \emph{predictive covert communication}, leverages predicted UAV locations to strategically control the transmit power of network nodes, thereby reducing the probability of detection by surveillance UAVs.
\item[\textbf{C2)}] We propose a data-driven methodology to achieve long-term predictions of UAV trajectories, which are inherently complex nonlinear dynamical systems. This approach integrates graph learning with Koopman theory within a Graph Neural Network (GNN) architecture, facilitating the learning of spatial interactions and linearization of dynamics in UAV networks.

\item[\textbf{C3)}] We conduct extensive simulations to validate our predictive model and compare its performance against several well-established baseline techniques. The results demonstrate superior accuracy in trajectory prediction and operational efficiency, thereby supporting the feasibility of low-latency covert operations in practical scenarios.
\end{itemize}

The remainder of the paper is structured as follows: in Section~\ref{sec:Sys}, we introduce our system model for a covert wireless ad-hoc network operating under the surveillance of multiple UAVs. Section~\ref{sec:ex} provides a brief review of existing methods for predicting UAV trajectories. Our proposed method for multi-UAV trajectory prediction and its corresponding training procedures are presented in Section~\ref{sec:gkae}. In Section~\ref{sec:SP}, we describe the simulation settings used to evaluate our approach. Section~\ref{sec:res} presents the results of our method and compares them against established baseline techniques. Finally, Section~\ref{sec:Conc} concludes the paper.

\begin{table}[h]
\centering
\caption{Glossary of Notions}
\label{tab:not}
\begin{tabular}{@{}>{\raggedright\arraybackslash}p{20mm}p{0.7\columnwidth}@{}} 
\toprule
Symbol & Description \\
\midrule
\textbf{UAV and Ground nodes} & \\
$\mathcal{L}$ & Set of indexed enemy UAVs \\
$\mathcal{N}_l(t)$ & Neighborhood of $l$-th UAV at time $t$ \\
$\bu_l(t)$ & Position vector of UAV $l$ at time $t$ \\
$\bw_n(t)$ & Position vector of ground node $n$ at time $t$ \\
$P_n(t)$ & Transmit power at ground node $n$ at time $t$ \\
$P_{l, n}(t)$ & Power received by UAV $l$ from ground node $n$ at $t$ \\
$\gamma_{i \to j}(t)$ & SNR at receiver $j$ from transmitter $i$ at time $t$ \\
\midrule
\textbf{GNN model} & \\
$\mathcal{V}$ & Set of nodes in the graph \\
$\mathcal{E}(t)$ & Dynamic edges in the graph at time $t$ \\
$\mathcal{X}(t)$ & Node feature matrix of UAVs at time $t$ \\
$\mathcal{A}(t)$ & Adjacency matrix of the graph at time $t$ \\
$\bh_G(t)$ & Graph embedding at time $t$ \\
$\bz_G(t)$ & Koopman invariant features at time $t$ \\
$\bK$ & Koopman matrix \\
\midrule
\textbf{Mathematical operations} & \\
$||\cdot||$ & Norm of a vector \\
$\Pr(\cdot)$ & Probability of an event \\
$\mathbb{E}[\cdot]$ & Expected value \\
$\circ$ & Element-wise multiplication \\
$\bigcup$ & Union operator \\
\bottomrule
\end{tabular}
\end{table}

\section{System Model and Problem formulation}\label{sec:Sys}

In this section, we formally define our system model for achieving covert wireless ad-hoc network against a multi-UAV surveillance network.   
\subsection{UAV Surveillance Network}
Let $L$ be the number of enemy UAVs performing surveillance which is represented by the set $\mathcal{L} = \{1, 2, \cdots, L\}$. We consider that the UAVs are dynamic in nature and its location at time $t$ is given using $\{{\bu_l(t): (u_{l_x}(t), u_{l_y}(t), u_{l_z}(t))}\}_{l \in \mathcal{L}}$. 

The dynamic model for the $l-$th UAV is given using 
\begin{align}
\mathbf{u}_l(t) = \mathbf{F}(\mathbf{u}_l(t-1)) + \mathbf{G}_m(\mathbf{u}_m(t-1))_{\forall m \in \mathcal{N}_l(t-1)},    
\end{align}
where $\mathcal{N}_l(t)$ represents the index of UAVs which is close to UAV $l$ and therefore will affect its dynamics. Let the neighborhood of UAV $l$ be calculated as
\be
\mathcal{N}_l(t) = \{m: ||\mathbf{u}_m(t) - \mathbf{u}_l(t)|| \le \tilde{D}\},
\ee
where $\tilde{D}$ is a distance threshold (in metres). 

An air-ground channel is assumed to be dominated by line-of-sight (LoS). The received signal strength at the $l$-th UAV from ground node $n$ at time $t$ is given as:
\be
P_{l, n}(t) = P_n(t) d_{l, n}(t)^{-\eta},
\ee
where $\eta$ is the path-loss exponent for air-ground channel.


\subsection{Terrestrial Ad-hoc Network}

We consider a terrestrial wireless ad-hoc network with $N$ ground nodes, denoted by the set $\mathcal{N} = \{1, 2, \cdots, N\}$ at  known locations given by $\{{\bw_n : (w_{n_x}, w_{n_y}, 0)}\}_{n \in \mathcal{N}}$. Here, $\bw_n \in \mathbb{R}^{3}$ represents the three-dimensional (3D) coordinates of the $n$-th ground node. Each ground node operates with an adjustable transmit power, which is represented by $P_n(t)$ at time $t$, $\forall n \in \mathcal{N}$. Then, the corresponding signal-to-noise ratio (SNR) at receiver $j$, denoted by $\gamma_{i \to j}$, becomes
\be
\gamma_{i \to j}(t) = \frac{P_i(t) d_{i, j}^{-\tilde \eta} \nu_{i,j} (t)}{N_0}, \forall i \ne j \in \mathcal{N} , 
    \label{EQ:SNR}
\ee
where $d_{i, j} = \cD(\bw_i, \bw_j) \ge 0$ is the large-scale fading term between the $i$-th and $j$-th nodes, $\nu_{i,j} (t) \ge$ 1 is the small-scale fading term, and $\tilde \eta$ and $N_0$ represent the path-loss exponent for ground communication and noise variance, respectively. 
Here, we assume that $\nu_{i,j} (t)$ is independent and identically distributed (iid), and $\cD(\bw_i, \bw_j) \ge ||\bw_i - \bw_j||$, as there might be obstacles between nodes $i$ and $j$, allowing the signal to travel a longer distance than the direct path, $||\bw_i - \bw_j||$. Thus, $\cD(\bw_i, \bw_j)$ can be considered as the effective distance between the two nodes.

To ensure a stable communication link between the ground nodes, the received SNR must exceed a predefined SNR threshold, represented by $\tilde{\gamma}$. Consequently, the set of communication links for node $i$ at time $t$ can be formally expressed as:
\be
\cM_i(t) = \{j : \gamma_{i \to j}(t) \ge \tilde{\gamma}\}_{i \ne j \in \mathcal{N}},
\ee
which is a random set due to the small-scale fading. We assume that there can be multiple communication rounds, say $T$. Then, the set of communication links can be re-defined as
\be 
\cM_i = \{j : \max_{t \in \{0,\ldots, T-1\}} \left\{ \gamma_{i \to j}(t)  \right\} \ge \tilde{\gamma}\}_{i \ne j \in \mathcal{N}} .
\ee 
If we assume that $\uE[\nu_{i,j} (t)] = 1$ for normalization purposes, it can be shown that 
\be 
\max_{t}  \gamma_{i \to j}(t)
\ge \frac{1}{T} \sum_{t=0}^{T-1} \gamma_{i \to j}(t)
\to \bar \gamma_{i \to j} = \frac{P_i(t) d_{i, j}  }{N_0},
\ee
based on the law of large numbers. Thus, with a sufficient large $T$, we can have the following subset of $\cM_i$:
\be 
\tilde \cM_i = \{j : \bar \gamma_{i \to j} \ge \tilde{\gamma}\}_{i \ne j \in \mathcal{N}}.
\ee 
Thus, it is possible to decide the transmit power $P_i (t) = P_i$ to ensure a sufficient number of links, say $|\tilde \cM_i| \ge \bar M > 0$, where $\bar M$ is the minimum number of links for each node. 

Note that no interference is considered in \eqref{EQ:SNR}. When there is interference, we can assume that the short-term fading term, $\nu_{i,j} (t)$, becomes small. In this case, $\nu_{i,j} (t)$ captures both the small-scale fading and interference. In addition, it is noteworthy that nodes do not continuously transmit, but they can only transmit when they have packets to send using a listen before talk (LBT) protocol.

\subsection{Power Control Problem for Covert Operations}

For the terrestrial ad-hoc network operating within a surveillance area of UAVs, controlling the transmit powers of nodes with respect to the locations of UAVs is necessary. To this end, we assume that $P_n$ is the nominal power of node $n$, used when no UAVs are in proximity. However, if UAVs are nearby, the transmit power must be lowered for covert operations.

Let
$w_n (t) = \max_{l} d_{l, n}^{-\eta}(t)$. The received power at UAV $l$ from ground node $n$, which is denoted by $P_{l, n}(t)$, has to be less than a given threshold, denoted by $\tilde{P}_{\text{det}}$. Consequently, we have
\be
P_{l, n}(t) = P_n d_{l, n}^{-\eta}(t) \le w_n (t) P_n\le \tilde{P}_{\text{det}},
\ee
From this, it can be shown that
\be 
P_n (t) = \min\left\{ P_n , \frac{\tilde{P}_{\text{det}}}{w_n (t)}
\right\} .
    \label{EQ:upp}
\ee 

There are several challenges involved in determining the power of node $n$ as outlined in \eqref{EQ:upp}. Ground nodes face difficulties in determining their power levels due to a lack of knowledge about the locations of UAVs. As previously mentioned, the CU can estimate the locations of UAVs, calculate the power level according to \eqref{EQ:upp}, and communicate this information to ground nodes. Given that fixed-wing UAVs typically fly at high speeds, any processing delay at the CU may result in the power control strategy in \eqref{EQ:upp} being based on outdated location information. Consequently, it becomes imperative for the CU to accurately predict the locations of UAVs in order to compensate for processing delays.

\section{Existing Methods for Predicting Trajectories of UAV}\label{sec:ex}
For covert wireless ad-hoc networks discussed in Section~\ref{sec:Sys}, it is necessary for the CU to predict the trajectories of surveillance UAVs hovering over the AO.

In this section, we present a few existing approaches for predicting UAV trajectories using machine learning techniques. Generally, UAVs are guided by a predefined \emph{UAV flight plan}. Thus, with sufficient historical data collected as a priori, it becomes feasible to learn the subsequent trajectory of a UAV. However, UAV trajectories usually exhibit nonlinear dynamics and depend on stochastic parameters such as weather conditions, making prediction a non-trivial task. We first review the works considering the prediction trajectory of a single UAV, and then discuss the case of multiple UAVs.

\subsection{Single UAV} 

\begin{table*}[t]
    \centering
    \caption{A literature review of the existing machine learning techniques of predicting UAV trajectories.}
    \begin{tabular}{p{1cm}p{1cm}p{6cm}p{1cm}p{1cm}p{6cm}}
        \toprule
        \textbf{Ref.} & \textbf{Approach} & \textbf{Key Insight} & \textbf{Pred. Input} & \textbf{Pred. Output} & \textbf{Challenges} \\
        \midrule
        \cite{Xue2017} & MLP & Utilizes MLP for trajectory prediction for small UAVs (s-UAV) circumventing the need for detailed control system knowledge or extensive aerodynamic parameter identification. & 2D loc., vel., wind, desired traj. & 1-step future loc. & \begin{enumerate}
            \item Simple MLP-based approach might provide limited temporal modelling leading to sub-optimal performance. 
            \item Wind and velocity parameters are considered as input, the data of which is often unavailable. 
        \end{enumerate} \\ 
        \midrule
        \cite{xiao2019trajectory} & RNN & Implements an RNN for real-time UAV trajectory prediction to optimize communication beamforming in smart cities. Utilizes angle-based data inputs reflecting UAV dynamics. & 2D angle data (elevation and horizontal) & 1-step future 2D angles & \begin{enumerate}
            \item Depends on Direction of Arrival (DOA) estimates to obtain 2D angle data, which could introduce processing delays, particularly when not aimed at long-term predictions.
            \item RNNs are often challenged by vanishing gradient issues resulting in unstable training. 
        \end{enumerate}\\
        \midrule
        \cite{shu2021trajectory} & Bi-LSTM & Utilizes bi-LSTM to predict UAV trajectory during cruise missions, enhancing real-time trajectory accuracy and ensuring safety through dynamic adjustments. & 3D loc. & 1-step future loc. & \multirow{4}{6cm}{\begin{enumerate}
            \item Long-term predictions are not explored.  
            \item LSTM-based approaches often lack expressiveness in capturing non-linear dynamics \cite{wang2017new}. 
        \end{enumerate}} \\
        \cmidrule(r){1-5}
        \cite{kant2022long} & LSTM-AE & Uses an AE-based LSTM network for better accuracy in predicting the UAV trajectory. & 3D loc., pressure, temperature & 1-step future loc. & \\
        \bottomrule
    \end{tabular}
    \label{tab:single}
\end{table*}

Deep neural networks (DNN)-based architectures, known for their universal approximation capabilities \cite{hornik1989multilayer, hornik1991approximation}, have been effectively employed for UAV trajectory prediction, as reviewed in Table~\ref{tab:single}. In \cite{Xue2017}, a multi-layer perceptron (MLP) is considered for predicting the trajectory of a UAV. However, it is well-established that MLPs yield suboptimal performance when handling sequential data characterized by governing dynamics \cite{lecun2015deep}. Other DNN-based architectures, such as recurrent neural networks (RNNs) \cite{rumelhart1985learning, jordan1997serial} and their variants, long short-term memory (LSTM) \cite{schmidhuber1997long} and gated recurrent units (GRU) \cite{cho2014learning}, emerge as superior alternatives for sequential data.

In \cite{xiao2019trajectory}, an RNN-based approach is used for trajectory prediction of UAVs in smart cities. Compared to MLPs, RNNs utilize recurrent connections, allowing for the processing of sequential data and the acquisition of patterns by preserving an internal memory. However, RNNs frequently encounter the vanishing gradient problem, which hampers the learning of long-term dependencies. To address this, LSTM introduces a gated-RNN architecture that maintains a stable gradient over time and retains information over longer sequences.

Other variants of the LSTM model, such as Bidirectional LSTM (Bi-LSTM), offer better performance than vanilla LSTM. For example, in \cite{shu2021trajectory}, Bi-LSTM is used for trajectory prediction of 3D UAV dynamics. Bi-LSTM captures temporal sequences from both past and future states, facilitating more effective prediction compared to vanilla LSTM models, which consider only past states.

Similarly, in \cite{kant2022long}, an LSTM-based autoencoder (LSTM-AE) is used to improve performance compared to the vanilla LSTM architecture. This model utilizes an encoder-decoder framework, with both components comprising LSTM networks. The encoder condenses the input trajectory data into a compact latent representation, while the decoder uses this condensed information to forecast future locations of the UAV trajectory. The encoder captures the temporal dynamics and key characteristics of the trajectory, while the decoder reconstructs or predicts the subsequent path that the UAV might follow.

The overarching limitations in the current literature are that predictions are typically made for a 1-step horizon, with long-term predictions not yet thoroughly investigated. To support real-time applications, it is crucial to develop techniques that facilitate predictions over extended horizons. Additionally, while LSTM is regarded as the state-of-the-art (SOTA) approach, it often falls short in accurately predicting non-linear dynamics \cite{wang2017new}.

Furthermore, in the development of intelligent autonomous systems \cite{zhou2020uav}, for tasks like surveillance \cite{xu2021communication}, the coordination of multiple UAVs is frequently taken into account, rather than relying solely on a single UAV. Therefore, predicting the trajectory of a multi-UAV network holds significant importance, akin to the necessity of predicting trajectories for single UAVs. However, as far as our understanding extends, there are presently no methods within the existing literature that explicitly consider this problem.
 
\subsection{Multiple UAVs} 

In this subsection, we briefly discuss modeling multi-UAV networks and the challenges involved in predicting their trajectories.

Bio-inspired swarm concepts can be applied to modeling multi-UAV networks \cite{xie2021bio, petravcek2020bio}. The fundamental idea behind biological swarms involves uniting individual entities, like birds, into a unified flock through local interactions. Similarly, in multi-UAV networks, each UAV autonomously adjusts its trajectory and actions based on local observations and interactions with neighboring UAVs. Several empirical models exist for capturing the dynamics of bio-inspired swarms, such as the Couzin model \cite{couzin2002collective} and Reynolds' Boid model \cite{reynolds1987flocks}, among others \cite{vicsek1995novel, cucker2007emergent}.

Predicting trajectories for multi-UAV networks presents additional challenges beyond those of single UAV trajectory prediction, including:
\begin{itemize}
\item Predicting trajectories sequentially for each of the $L$ UAVs in the network requires implementing $L$ separate predictive models. This approach is impractical due to the substantial computational burden of training $L$ models, each tailored to a single UAV.
\item Alternatively, employing a unified predictive architecture for all UAVs in the network leads to a linear increase in input size proportional to $L$. This scenario also results in significant computational demands, as it requires more layers and a higher number of hidden nodes in the predictive architecture to manage the growing complexity and input volume.
\end{itemize}





\

\section{GKAE and Its Application to Predictive Covert
Communication}\label{sec:gkae}

In this section, we propose a novel approach to modeling a multi-UAV network based on a graph representation and a neural network that can capture the dynamics of UAVs in the network.

\begin{figure}[h!]
    \centering
\includegraphics[width=1 \columnwidth]{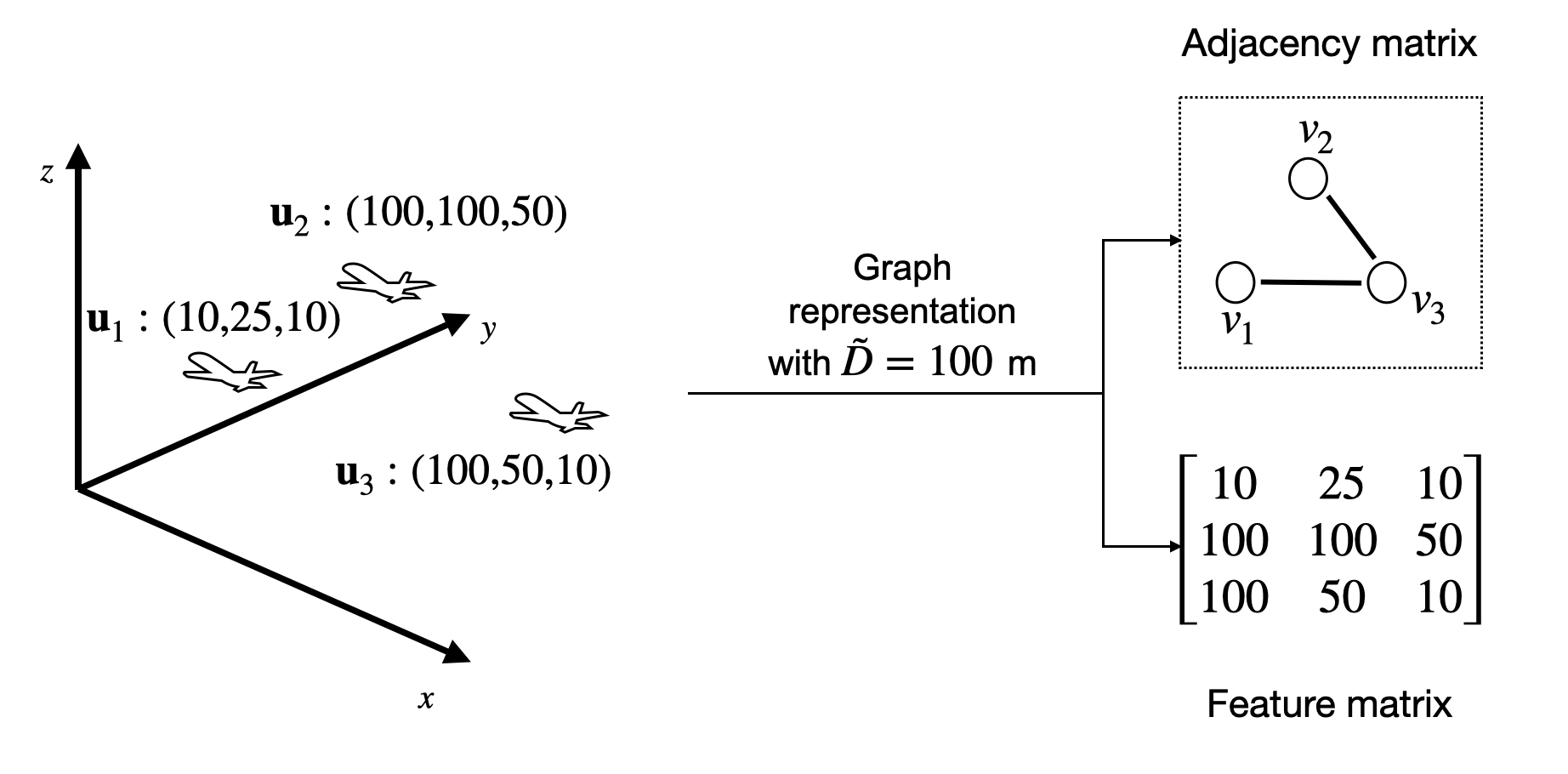}
    \caption{
A schematic graph representation is employed for a multi-UAV network, utilizing node feature and adjacency matrices, where UAVs within 100 meters are connected.} 
    \label{fig:grep}
\end{figure}

The multi-UAV network can be effectively modeled using a graph, which allows us to represent not only the information related to each UAV but also the spatial dependencies through the edges. In addition, due to the mobility of UAVs, we define a time-varying graph, whose representation at time $t$ is given by $G(t) = (\mathcal{V}, \mathcal{E}(t), \mathcal{X}(t), \mathcal{A}(t))$, where 
\begin{itemize}
    \item $\mathcal{V} = \{v_1, \cdots, v_L\}$ is the set of nodes.
    \item $\mathcal{E}(t) = \{e_{ij} (t)\}$ is the set of edges, with $e_{ij} (t) = 1$ if $|\bu_i (t) - \bu_j (t)| \le \tilde{D}$, $\forall i, j \in \mathcal{V}$; otherwise $e_{ij} (t) = 0$.
    \item $\mathcal{X}(t) = [\mathbf{u}_1 (t) \  \mathbf{u}_2 (t) \ldots \mathbf{u}_L (t)] \in \mathbb{R}^{L \times 3}$ is the node feature matrix. That is, each UAVs' feature is its 3-dimensional location.
    \item $\mathcal{A}(t) \in \mathbb{R}^{L \times L}$ is the adjacency matrix such that the entry $A_{ij}(t) = 1$ if $e_{ij} (t) = 1$, and $A_{ij} (t) = 0$ otherwise. 
    \end{itemize}
In Fig.~\ref{fig:grep}, we represent a 3-dimensional UAV network using a graph with its set of nodes, edges, node features, and adjacency matrix.

\subsection{Graph Neural Network} 

Graph neural networks (GNNs) extend traditional DNNs to graph data by generalizing the convolution operation to learn spatial representations \cite{hamilton2020graph}. GNNs utilize a form of \emph{neural message passing}, where vector messages are exchanged between nodes and updated using neural networks. This process includes:
\begin{itemize}
    \item \textit{Aggregation phase}: Each node gathers and aggregates information from its neighboring nodes at time $t$. 
    \item \textit{Updating phase}: After aggregating information, each node updates its feature representation based on the aggregated information at time $t$
\end{itemize} 
Mathematically, as summarized in \cite{xu2018powerful}, the message passing to calculate the node embedding for the $\ell$-th layer in the GNN for the $i$-th node at time $t$ can be described as:
\begin{align}
    \textbf{AGG}(\cdot) \quad &\mathbf{a}_{i}^{(\ell)}(t) = \mathcal{A}\left(\{\mathbf{h}_{j}^{(\ell-1)}(t) : v_j \in \mathcal{N}(v_i)\}\right) \\
    \textbf{UPD}(\cdot) \quad &\mathbf{h}_{i}^{(\ell)}(t) = \mathcal{U}\left(\mathbf{h}_{i}^{(\ell-1)}(t), \mathbf{a}_{i}^{(\ell)}(t) \right),
\end{align}
where $\mathbf{h}^{(0)}_{i}(t) = \mathbf{x}_{i}(t)$, and $\mathcal{A}(\cdot)$ and $\mathcal{U}(\cdot)$ represent the aggregation and update functions, respectively. The choice of the aggregation and update functions in GNNs is crucial and heavily reliant on the nature of the problem \cite{hamilton2020graph}. Through message passing, GNNs can decode the \emph{structural} information via the adjacency matrix and describe \emph{feature-based} information through the local aggregation of node features which is analogous to the convolution filters in convolution neural networks (CNN). 
The core message passing operation occurs at the node level, while some tasks might require a coarse representation of the graph, which is typically achieved using a readout function. The readout function performs a permutation-invariant aggregation on the node embeddings as follows:
\be
\mathbf{h}_{G} = \textbf{READ}\left(\{\mathbf{h}_{i}^{(\ell)} : v_i \in \mathcal{V}\}\right),
\ee
where \textbf{READ}($\cdot$) can be a simple permutation invariant function such as summation or a more sophisticated function \cite{buterez2022graph, navarin2019universal}.

\subsection{Koopman AutoEncoder}

While GNNs in our architecture are used for learning compact graph embeddings, additional modeling is required to capture the dynamics for long-term predictions. To this end, the Koopman operator theory \cite{Koopman31} \cite{Lasota98} offers a promising alternative to existing methods like LSTM and RNN. Prior to detailing our proposed approach, we briefly introduce the Koopman operator theory to lay groundwork for learning dynamics over graph embeddings. 

\begin{itemize}
    \item \textbf{Koopman operator theory} \cite{Brunton22}: Consider a state vector $\bx(t)\in \cX \subseteq \uR^N $ of a nonlinear dynamical system at  (discretized)  time $t \in \uN$, where $\cX$ stands for the state space. The nonlinear  dynamics of $\bx(t)$ without external inputs can be described as
\be 
\bx(t+1) = \bF(\bx(t)),
    \label{EQ2:xFx}
\ee 
where $\bF: \cX \to \cX$ is a flow map.
Consider that $g: \cX \to \uR$ is a measurement function, which is a function of state $\bx(t)$ and called an observable. Koopman operator theory states that there exists a linear (infinite-dimensional) operator $\cK$ that acts to advance $g$, i.e.,
\begin{align}
\cK g = g \circ \bF,
    \label{EQ3:KO}
\end{align}
where $\circ$ represents the composition operator.
Applying \eqref{EQ3:KO} to \eqref{EQ2:xFx}, we have
\begin{align}
g(\bx(t+1)) = g \circ \bF(\bx(t)) = \cK g(\bx(t)),
    \label{EQ:gg}
\end{align}
where $g(\bx(t))$ is the observable measured at time $t$.
This can be extended to cases with multiple observables. Precisely, let $\bg (t) = [g_1 (t) \ \ldots \ g_M (t)]^\rT$, where $g_m (t) = g_m (\bx(t))$, which is called the observable vector. 
Then, from \eqref{EQ:gg}, we have
\be 
\bg(t+1) = \cK \bg(t) \in \uR^M. 
\ee 

If $\bg(t) \in \cG$ and $\cK \bg(t) \in \cG$, where $\cG$ is a finite-dimensional space, $\cG$ becomes a Koopman invariant subspace  \cite{Brunton_PLOS}. In this case, $\cK$ becomes a finite-dimensional linear operator and is represented by a matrix $\bK \in \uR^{M \times M}$, which is called the Koopman matrix of dimension $M$. The eigenvalues and eigenvectors of $\bK$ can describe the linear evolution of the dynamical system in the Koopman invariant subspace. A prerequisite to discovering $\bK$ is to find a Koopman invariant subspace. 
Thanks to data-driven approaches, this subspace can be identified using numerical techniques with predefined kernel functions \cite{Mauroy2020-es} \cite{Brunton22}. With the success of deep learning based approaches, recent frameworks rely on training an encoder-decoder structured deep neural networks \cite{Takeishi17} \cite{lusch2018deep} \cite{azencot2020forecasting}. 
\end{itemize}

\subsection{The Proposed Framework: GKAE for Multi-UAV}

The approach in \cite{lusch2018deep} adopts the autoencoder architecture to model the nonlinear dynamics, where the encoder produces the observable vector , $\bg(t)$, and the decoder converts $\bg(t+1)$ to $\bx(t+1)$. As a result, this approach is referred to as the Koopman Autoencoder (KAE). Since the KAE is applicable to dynamics in the Euclidean space, it can be used to model the dynamics of a single UAV. However, when it is applied to multiple interacting UAVs in a network, it is necessary to consider a graph representation to capture their interactions. In this subsection, we aim to modify the KAE so that it can model the dynamics of multi-UAV networks.

\begin{figure*}[t]
    \centering
    \includegraphics[width=0.9\textwidth]{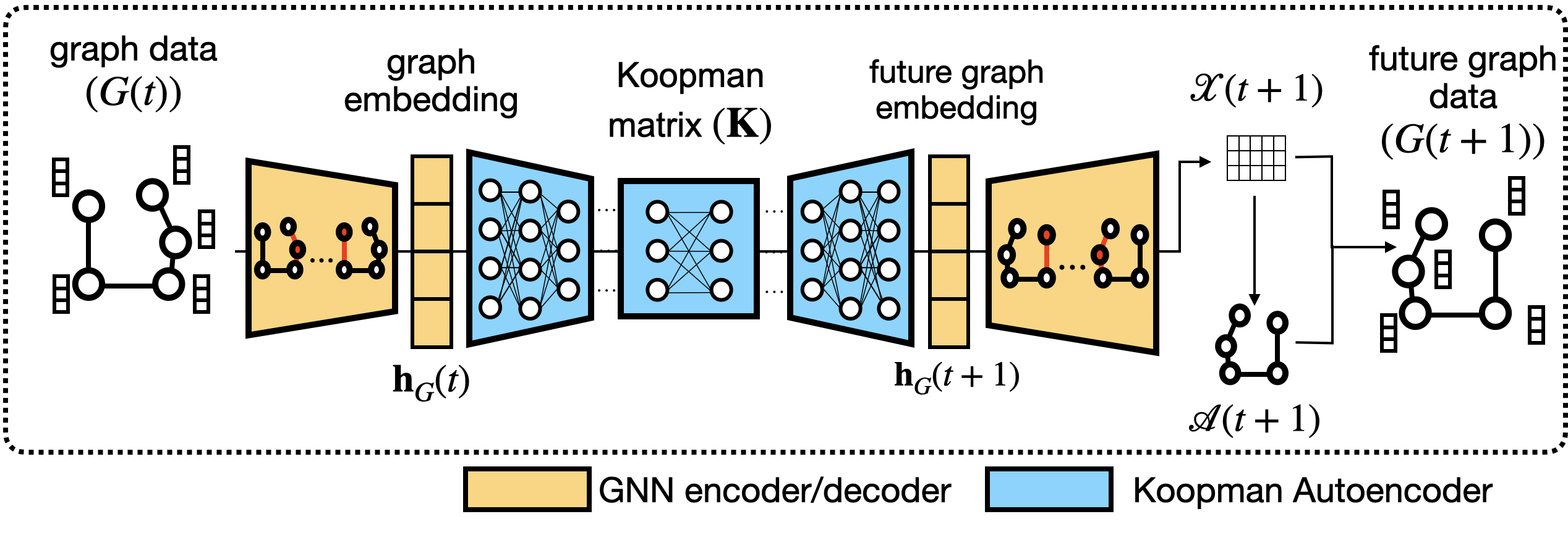}
    \caption{An illustration of the proposed architecture: Graph-based Koopman Autoencoder (GKAE).} 
    \label{fig:prop}
\end{figure*}

We dub our framework the \emph{Graph KAE (GKAE)} as it integrates GNN with the KAE, which is illustrated in Fig.~\ref{fig:prop}. The GKAE employs an autoencoder structure based on the KAE with several trainable components. The roles of the key components are stated as follows:
\begin{itemize}
\item \textbf{GNN encoder}: The input to the graph encoder for message-passing is the graph $G(t)$, and the output is a coarse graph embedding used for feature representation of the graph.
\item \textbf{Koopman encoder}: Utilizing the coarse graph representation as the state vector, the DNN encoder performs the lifting operation to represent the graph embedding in the Koopman invariant subspace. This step aims to linearize the dynamics of the graph.
\item \textbf{Koopman matrix}: The Koopman matrix is employed to facilitate linear predictions within the Koopman invariant subspace.
\item \textbf{Koopman decoder}: The DNN decoder reconstructs the predictions made in the Koopman invariant subspace back to the original coarse graph embedding representations.
\item \textbf{GNN decoder}: Leveraging the decoded coarse graph embeddings, the GNN decoder reconstructs the graph structure, resulting in a predicted graph.
\end{itemize}

The operations of the key components can be summarized as follows:
\begin{align}
\label{EQ:G1}
    \mathbf{h}_G(t) &=\mathcal{E}_\text{g}(G(t); \boldsymbol{\varphi}) \\
\label{EQ:G2}
    \mathbf{z}_G(t) &= \mathcal{E}_\text{d}(\mathbf{h}_G(t); \boldsymbol{\theta})\\
\label{EQ:G3}
    \mathbf{z}_G(t+1) &= \mathbf{K}\mathbf{z}_G(t)\\
\label{EQ:G4}    
    \mathbf{h}_G(t+1) &= \mathcal{D}_{\text{d}}(\mathbf{z}_G(t+1); \boldsymbol{\theta}^{-1})\\
\label{EQ:G5}    
    \mathcal{G}(t+1) &= \mathcal{D}_{\text{g}}(\mathbf{h}_G(t+1); \boldsymbol{\varphi}^{-1}).
\end{align}
In \eqref{EQ:G1}, the graph encoder converts the graph sample at time $t$ to its embedding, where $\boldsymbol{\varphi}$ represents the parameter vector of the graph encoder. The output of the graph encoder becomes the input to the Koopman encoder as shown in \eqref{EQ:G2}, where $\boldsymbol{\theta}$ stands for the parameter vector of the Koopman encoder. In \eqref{EQ:G3}, the observable vector advances through the Koopman matrix in the Koopman invariant subspace. In \eqref{EQ:G4}, the Koopman decoder reconstructs the predictions made in the Koopman invariant subspace. Finally, in \eqref{EQ:G5}, the GNN decoder leverages the decoded embeddings to reconstruct the graph structure, resulting in the predicted graph at time $t+1$. Here, $\boldsymbol{\theta}^{-1}$ and $\boldsymbol{\varphi}^{-1}$ represent the parameter vectors of the Koopman and graph decoders, respectively. Each of encoders and decoders is implemented by DNN. Thus, there are five parameters to be trained: $\boldsymbol{\varphi}$ (of DNN), $\boldsymbol{\theta}$ (of DNN), $\bK$ (of a square matrix), $\boldsymbol{\theta}^{-1}$ (of DNN), and $\boldsymbol{\varphi}^{-1}$ (of DNN). Note that \eqref{EQ:G2} - \eqref{EQ:G4} are the operations of the KAE as described in \cite{lusch2018deep}, while \eqref{EQ:G1} and \eqref{EQ:G5} pertain to the embedding and subsequent reconstruction of the graph.

In summary, using the GKAE, the locations of multiple interactive UAVs can be predicted as follows:
\begin{eqnarray} 
&  G(t)\  {\buildrel \eqref{EQ:G1} \over \to}\ \bh_G (t) \ {\buildrel \eqref{EQ:G2} \over \to} \  \bz_G(t) & \cr 
&  {\buildrel \eqref{EQ:G3} \over \to} \  \bz_G (t+1) \  {\buildrel \eqref{EQ:G4} \over \to} \ \bh_G (t+1) \  {\buildrel \eqref{EQ:G5} \over \to} \
G(t+1). &
\end{eqnarray}
That is, by taking the feature matrix $\cX(t+1)$ of $G(t+1)$, the locations of the UAVs at time $t+1$ can be obtained. To this end, the encoders, decoders, and the Koopman matrix need to be trained, which will be discussed in the next subsection.

\subsection{Training}
We adopt a two-step training process for our proposed method, motivated by the need to effectively capture and predict the dynamics of a multi-UAV network. 

Firstly, we focus on training the graph encoder and graph decoder using the graph data from the multi-UAV network. The primary objective in this step is to accurately reconstruct the graph structure based on the graph embedding generated by the graph encoder. This reconstruction task ensures that the encoder captures the essential features and relationships within the graph, while the decoder learns to map these embeddings back to the original graph structure. In the second step, we utilize the sequential graph embeddings, obtained from the first step, as inputs to the KAE. The KAE comprises three main components: the Koopman encoder, the Koopman matrix, and the Koopman decoder. This model is designed to learn the underlying dynamics within the graph embeddings in a linear manner, leveraging the Koopman operator theory. By doing so, the KAE enables linear prediction of future graph embeddings, effectively capturing the temporal evolution of the multi-UAV network.

Considering a graph dataset of multi-UAV trajectories denoted by $\{G(t)\}_{t=1}^T$, we define the graph reconstruction loss as follows:
\begin{equation}
L_{\text{grec}} = \sum_{t=1}^T \left\| G(t) - \mathcal{D}_{\text{g}}(\mathcal{E}_{\text{g}}(G(t))) \right\|^2,
\end{equation}
where our focus is mainly on the node feature matrix corresponding to the locations of the UAVs. Given the reconstructed node feature matrix $\mathcal{X}(t)$, the structure of the graph can be reconstructed using a pre-defined neighborhood rule, ensuring that the spatial relationships and connectivity between the UAVs are preserved.
While training the KAE, we consider the following loss terms:
\begin{align}
    L_{\text{rec}} &= \sum_{t=1}^T \left\| \mathbf{h}_G(t) - \mathcal{D}_{\text{d}}(\mathcal{E}_{\text{d}}(\mathbf{h}_G(t))) \right\|^2, \\
    L_{\text{pred}} &= \sum_{\Delta t=1}^\tau \left\| \mathbf{h}_G(t + \Delta t) - \mathcal{D}_{\text{d}}(\mathbf{K}^{(\Delta t)} \mathbf{z}_G(t)) \right\|^2,
\end{align}
where $L_{\text{rec}}$ is the reconstruction loss, which focuses on reconstructing the graph embeddings, and $L_{\text{pred}}$ is the forward prediction loss, which compares the predicted future embeddings with the original ones. In the forward prediction loss, $\tau$ is the hyperparameter that defines the linear horizon during training.
The overall optimization for training the GKAE can be written as follows: 
\be
\min_{\boldsymbol{\varphi}, \boldsymbol{\theta}, \mathbf{K}, \boldsymbol{\varphi}^{-1}, \boldsymbol{\theta}^{-1}} \alpha_1(L_{\text{grec}}) + \alpha_2(L_{\text{rec}} + L_{\text{pred}}), 
\ee
where $\alpha_1$ and $\alpha_2$ are weight parameters for the corresponding loss terms.

\section{Simulation Parameters}\label{sec:SP}
\subsection{Multi-UAV Trajectory Data}
In this section, we describe the UAV trajectory modeled using the Couzin dynamics framework \cite{couzin2002collective}, which has been further elaborated using in Algorithm~\ref{Algorithm:dynamics}. The Couzin dynamics framework delineates three distinct interaction zones among the UAVs, described by
\begin{itemize}
    \item Zone of repulsion ($r_{\text{rep}}$): This parameter ensures that UAVs avoid collisions by maintaining a minimum distance from one another.
    \item Zone of alignment ($r_{\text{ali}}$): This parameter facilitates the synchronization of UAV direction and speed with the average direction of neighboring UAVs. 
    \item Zone of attraction ($r_{\text{att}}$): This parameter promotes cohesive swarm behavior by causing UAVs to move towards one another. 
\end{itemize} 
We consider an area of operation of $500 \times 500 \text{m}^2$ where multiple UAVs are deployed for surveillance at regular short intervals. Given the limitations in battery power, it is assumed that the UAVs can only provide surveillance for a short duration. Therefore, to cover the entire operational area in this limited time, multiple UAVs are utilized. 

Furthermore, for an effective surveillance operation, it is essential that UAVs achieve maximum area coverage while minimizing overlap in their coverage areas. This can be accomplished by maximizing the zone of attraction to foster swarm-like behavior, while effectively maintaining a zone of repulsion between the UAVs to ensure they are distributed across different areas for surveillance. Unless specified otherwise, the simulation parameters for the multi-UAV dynamics are given in Table.~\ref{tab:combinedParameters}.
\begin{algorithm}
\caption{UAV Dynamics Simulation} \label{Algorithm:dynamics}
\begin{algorithmic}[1]
\State \textbf{Inputs:} $L, V_{\text{max}}, r_{\text{rep}}, r_{\text{ori}}, r_{\text{att}}, \Delta t, \theta_{\text{max}}, Z_{\text{min}}, Z_{\text{max}}, X_{\text{size}}$

\State \textbf{Initialize:}
\State $\mathbf{u}_i(0) \gets$ Random positions within $[0, X_{\text{size}}]^3$
\State $\mathbf{u}_i(0)_z \gets$ Random altitudes in $[Z_{\text{min}}, Z_{\text{max}}]$
\State $\mathbf{v}_i(0) \gets$ Random unit vectors scaled by $V_{\text{max}}$

\For{$t = 0$ to $T$ step $\Delta t$}
    \For{each UAV $i$}
        \State Initialize $\mathbf{f}_{\text{rep}}(i) \gets \mathbf{0}$, $\mathbf{f}_{\text{ori}}(i) \gets \mathbf{0}$, $\mathbf{f}_{\text{att}}(i) \gets \mathbf{0}$
        \For{each UAV $j \neq i$}
            \If{$d_{ij} < r_{\text{rep}}$}
                \State $\mathbf{f}_{\text{rep}}(i) \gets \mathbf{f}_{\text{rep}}(i) + d_{ij}$
            \ElsIf{$d_{ij} < r_{\text{ori}}$}
                \State $\mathbf{f}_{\text{ori}}(i) \gets \mathbf{f}_{\text{ori}}(i) + \mathbf{v}_j(t)$
            \ElsIf{$d_{ij} < r_{\text{att}}$}
                \State $\mathbf{f}_{\text{att}}(i) \gets \mathbf{f}_{\text{att}}(i) - d_{ij}$
            \EndIf
        \EndFor
        \State $\mathbf{v}_{\text{des}}(i) \gets \mathbf{v}_i(t) + \mathbf{f}_{\text{rep}}(i) + \mathbf{f}_{\text{ori}}(i) + \mathbf{f}_{\text{att}}(i)$
        \State $\mathbf{v}_{\text{des}}(i) \gets \text{limit\_speed}(\mathbf{v}_{\text{des}}(i), V_{\text{max}})$
        \State $\mathbf{v}_i(t + \Delta t) \gets \text{limit\_turning}(\mathbf{v}_i(t), \mathbf{v}_{\text{des}}(i), \theta_{\text{max}})$
        \State $\mathbf{u}_i(t + \Delta t) \gets \mathbf{u}_i(t) + \mathbf{v}_i(t + \Delta t) \times \Delta t$
        \State $\mathbf{u}_i(t + \Delta t)_z \gets \max(\min(\mathbf{u}_i(t + \Delta t)_z, Z_{\text{max}}), Z_{\text{min}})$
    \EndFor
\EndFor

\State \textbf{Functions:}
\State \textbf{function} limit\_speed($\mathbf{v}$, $V_{\text{max}}$)
\State \ \ \ Return $\min(\|\mathbf{v}\|, V_{\text{max}}) \times \frac{\mathbf{v}}{\|\mathbf{v}\|}$
\State \textbf{function} limit\_turning($\mathbf{v}_{\text{current}}$, $\mathbf{v}_{\text{desired}}$, $\theta_{\text{max}}$)
\State \ \ \ Calculate $\phi_{\text{current}} = \arctan2(\mathbf{v}_{\text{current}}[1], \mathbf{v}_{\text{current}}[0])$
\State \ \ \ Calculate $\phi_{\text{desired}} = \arctan2(\mathbf{v}_{\text{desired}}[1], \mathbf{v}_{\text{desired}}[0])$
\State \ \ \ $\Delta\phi = \phi_{\text{desired}} - \phi_{\text{current}}$
\State \ \ \ If $\Delta\phi > \theta_{\text{max}}$, then $\Delta\phi = \theta_{\text{max}}$
\State \ \ \ If $\Delta\phi < -\theta_{\text{max}}$, then $\Delta\phi = -\theta_{\text{max}}$
\State \ \ \ Return $V_{\text{max}} \times [\cos(\phi_{\text{current}} + \Delta\phi), \sin(\phi_{\text{current}} + \Delta\phi), 0]$
\end{algorithmic}
\end{algorithm}
\begin{algorithm}
\caption{Transmit Power Calculation for Covert Ground Nodes} \label{algorithm:calc}
\begin{algorithmic}[1]
\State \textbf{Inputs:} 
\State $L$, $N$, $\tilde{P}_{\text{det}}$, $\eta$, $P_{\text{max}}$, $\mathbf{u}_l(t)$, $\mathbf{w}_n$
\State \textbf{Initialize:}
\State $P_n(t) \gets P_{\text{max}}$ for all $n \in \mathcal{N}$

\Procedure{Calculate Transmit Power}{}
    \For{each ground node $n$}
        \State $\text{max\_dist\_inv}(t) \gets 0$
        \For{each UAV $l$}
            \State $d_{l, n}(t) \gets ||\mathbf{u}_l(t) - \mathbf{w}_n||$
            \State $\text{dist\_inv}(t) \gets d_{l, n}(t)^{-\eta}$
            \If{$\text{dist\_inv}(t) > \text{max\_dist\_inv}(t)$}
                \State $\text{max\_dist\_inv}(t) \gets \text{dist\_inv}(t)$
            \EndIf
        \EndFor
        \State $w_n(t) \gets \text{max\_dist\_inv}(t)$
        \State $P_n(t) \gets \min(P_{\text{max}}, \tilde{P}_{\text{det}} / w_n(t))$
    \EndFor
\EndProcedure
\end{algorithmic}
\end{algorithm}
\begin{table}[t]
\centering
\caption{Simulation Parameters for the Environment and UAV Dynamics.}
\begin{tabular}{ll}
\toprule
Parameter & Value \\
\midrule
\textbf{Environment Parameters} & \\
Area of operation & $500 \times 500 \ \text{m}^2$ \\
Num. of UAVs $(L)$ & 4 \\
Number of Ground Nodes ($N$) & 25 \\
Maximum Transmit Power (\(P_{\text{max}}\)) & 20 W \\
Threshold for Received Power (\(\tilde{P}_{\text{det}}\)) & 1 \(\mu \text{W}\) \\
Path-loss constant for air-ground channel ($\eta$) & 1 \\
Noise variance ($N_0$) & -174 dBm/Hz \\
Descaling factor ($\lambda$) & 0.5 \\
Prediction horizon ($H$) & 10 sec. \\
Weight parameters ($\alpha_1, \alpha_2$) & 1, 1 \\
\midrule
\textbf{UAV Dynamics Parameters} & \\
Area of operation & $500 \times 500 \ \text{m}^2$ \\
Time step ($\Delta t$) & $0.1\ \text{s}$ \\
Num. of UAVs $(L)$ & 4 \\
Maximum speed ($V_\text{max}$) & $20 \text{m/s}$ \\
Maximum turning angle ($\theta_{\text{max}}$) & $\frac{\pi}{100}$ \ \text{rad.} \\
Zone of repulsion ($r_{\text{rep}}$) & 300 \ \text{m} \\
Zone of alignment ($r_{\text{ali}}$) & 0 \ \text{m} \\
Zone of attraction ($r_{\text{att}}$) & 500 \ \text{m} \\
\bottomrule
\end{tabular}
\label{tab:combinedParameters}
\end{table}

\subsubsection{Architecture Details of the GKAE}
Our proposed method comprises a total of 13 layers. Initially, to learn the structure of our graphs, we employ two SAGE convolution layers, each with 4 hidden neurons. The KAE is defined with seven fully connected layers: the Koopman encoder and decoder each contain three layers with 16 hidden neurons. The Koopman matrix is represented by a fully connected layer with 8 hidden neurons. Finally, the graph decoder includes four fully connected layers with configurations of {4, 4, 4, {2, 3}}, where the number of neurons in the last layer is contingent on the dimensionality of the dataset, specifically whether it is 2-dimensional or 3-dimensional.

The GKAE is trained for 400 epochs using an NVIDIA RTX 6000 Ada generation GPU. The exponential linear unit (ELU) activation function is employed in the graph encoder and graph decoder, while the hyperbolic tangent (tanh) activation function is used in the Koopman encoder and Koopman decoder. The Adam optimizer is used for optimization of our proposed model.
\subsubsection{Baselines} We consider our predictions with several well established methods, which are used as baselines for performance evaluation
\begin{itemize}
\item[\textbf{B1})] \textbf{Recurrent Neural Network (RNN):} An RNN with 10 layers and 16 hidden neurons, configured for a sequence length of 0.8 seconds, to predict the next sequences. 
\item[\textbf{B2})] \textbf{Gated Recurrent Units (GRU):} A GRU model with 10 layers and 16 hidden neurons, also configured for a sequence length of 0.8 seconds length. 
\item[\textbf{B3})] \textbf{Long Short-Term Memory (LSTM):} Three variants are considered: 
\begin{itemize}
    \item[\textbf{B3.1})]Vanilla LSTM.
    \item[\textbf{B3.2})] Bidirectional LSTM.
    \item[\textbf{B3.3})] Autoencoder-based LSTM (LSTM-AE).
\end{itemize}. 
Each of the methods is defined with 10 layers with similar input configurations. 
\end{itemize}
\subsection{Performance metric}
For evaluating the methods, we make future predictions when only the initial locations of the UAVs are specified, $\bu_i(t), i \in \mathcal{L}$. For a single prediction step $\Delta t$, the prediction error over $L$ predictions of the UAV locations is given as the squared difference between the true location, $\bu({t + \Delta t})$ and the predicted locations, $\hat{\mathbf{u}}({t + \Delta t})$, given as:
\be
\epsilon_{\text{pred}}(t + \Delta t) = \frac{1}{L} \sum_{i=1}^{L} \| \mathbf{u}_i(t + \Delta t) - \hat{\mathbf{u}}_i(t + \Delta t) \|^2.
\ee
Then, over \(\Delta t = 1, \cdots, H\) seconds of predictions, we have the average prediction error as:
\be
\epsilon_{\text{mean}} = \frac{1}{H} \sum_{\Delta t = 1}^H\epsilon_{\text{pred}}(t + \Delta t).
\ee
\subsection{Predictive Covert Communication}
For predictive covert communication, we require the accurate predicted trajectories of the $L$ UAVs for $\Delta t = 1, \cdots H$ seconds, where the ground nodes can do covert communication with the predicted locations of the UAV trajectories. Using this information of the predicted trajectories, we can define the upper bound transmit power for the ground nodes, as seen in Algorithm~\ref{algorithm:calc}. The upper bound establishes the transmit power which can maximize the data rate of a ground node while ensuring no detection from the UAVs. 

\begin{figure}[h!]
    \centering
\includegraphics[width=0.9 \columnwidth]{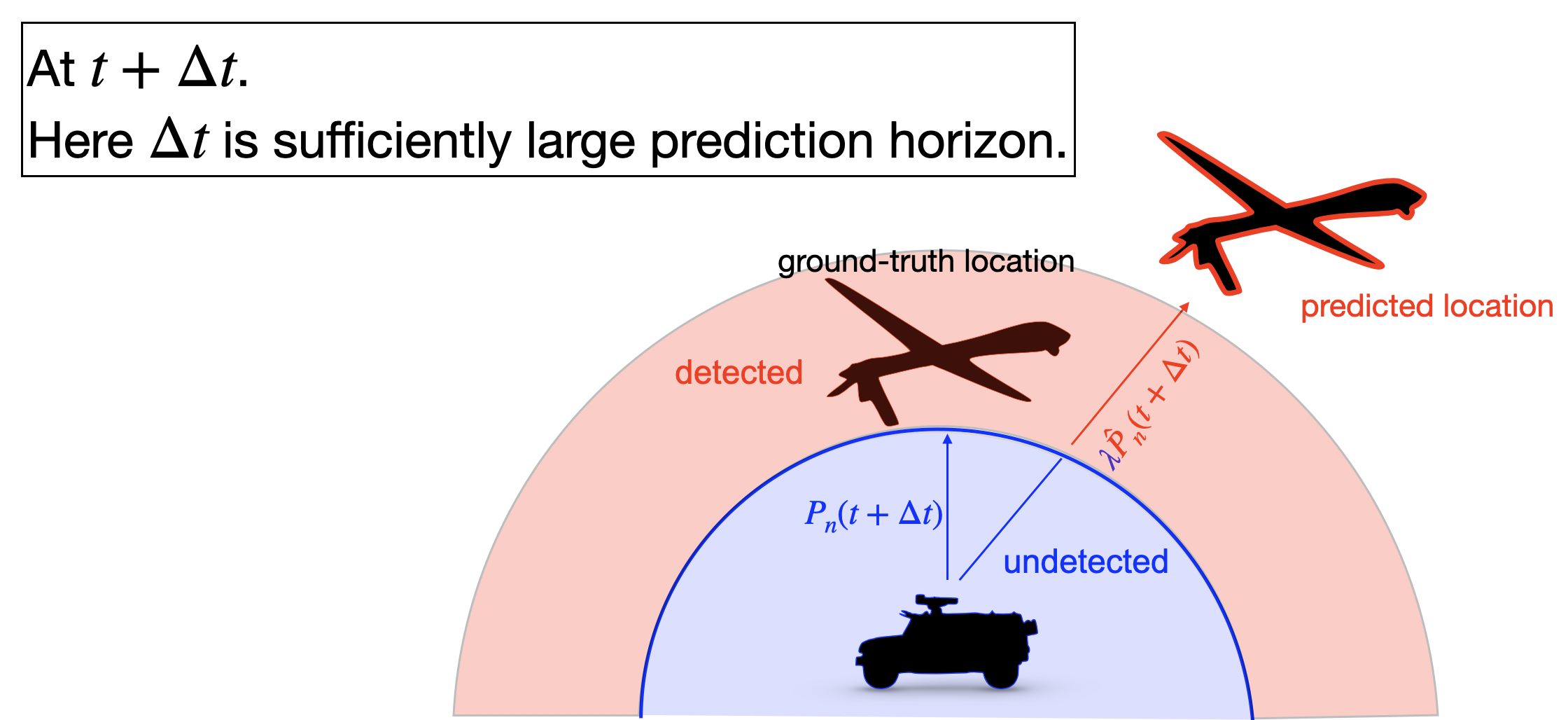}
    \caption{Disparity in predicted location versus true location can lead to detection. Downscaling the transmit power helps in preventing detection events.} 
    \label{fig:det}
\end{figure}

However, when predicting on a longer horizon (in sec.), we introduce the prediction error in the trajectories, which might lead to incorrect estimations of the upper bound of the transmit power of the UAVs, leading to detection, as illustrated in Fig.~\ref{fig:det}.
To account for prediction errors, we introduce a descaling factor $\lambda$, where $0 < \lambda < 1$, which adjusts the estimated transmit power based on the predicted UAV locations. For a prediction at $\Delta t$ second, the transmit power for node $n$ using the predicted model is denoted as $\hat{P}_n(t + \Delta t)$, while the transmit power using the actual location of the UAVs is denoted as $P_n(t + \Delta t)$. 
\begin{table*}[h]
\centering
\caption{Comparison of the training settings and predictive performance for the various methods.}
\label{tab:training_parameters}
\begin{tabular}{p{2cm}p{2cm}p{2cm}p{3.8cm}p{3.8cm}}
\toprule
\multicolumn{1}{c}{} & \multicolumn{2}{c}{\textbf{Training}} & \multicolumn{2}{c}{\textbf{Prediction}} \\
\cmidrule(r){2-3} \cmidrule(l){4-5}
\textbf{Method} & \textbf{Epochs} & \textbf{Params.} & \textbf{Pred. error (2D)} & \textbf{Pred. error (3D)} \\
\midrule
GKAE & 400 & 2K & \textbf{0.0011$\pm$0.0001} \hspace{1mm}
\begin{tikzpicture}[baseline=-0.75ex,xscale=10]
    \draw[thick] (0,0) -- (0.1,0);
    \filldraw[fill=white] (0.0011-0.0001,-0.1) rectangle (0.0011+0.0001,0.1);
    \draw[thick] (0.0011,-0.1) -- (0.0011,0.1);
\end{tikzpicture} & \textbf{0.0050$\pm$0.0012} \hspace{1mm}
\begin{tikzpicture}[baseline=-0.75ex,xscale=3]
    \draw[thick] (0,0) -- (0.3,0); 
    \filldraw[fill=white] (0.0050-0.0012,-0.1) rectangle (0.0050+0.0012,0.1);
    \draw[thick] (0.0050,-0.1) -- (0.0050,0.1);
\end{tikzpicture} \\ 
LSTM-AE & 508 & 21K & 0.0112$\pm$0.0018 \hspace{1mm}
\begin{tikzpicture}[baseline=-0.75ex,xscale=10]
    \draw[thick] (0,0) -- (0.1,0); 
    \filldraw[fill=white] (0.0112-0.0018,-0.1) rectangle (0.0112+0.0018,0.1);
    \draw[thick] (0.0112,-0.1) -- (0.0112,0.1);
\end{tikzpicture} & 0.0324$\pm$0.0049 \hspace{1mm}
\begin{tikzpicture}[baseline=-0.75ex,xscale=3]
    \draw[thick] (0,0) -- (0.3,0); 
    \filldraw[fill=white] (0.0324-0.0049,-0.1) rectangle (0.0324+0.0049,0.1);
    \draw[thick] (0.0324,-0.1) -- (0.0324,0.1);
\end{tikzpicture} \\ 
GRU & 174 & 6K & 0.0601$\pm$0.0071 \hspace{1mm}
\begin{tikzpicture}[baseline=-0.75ex,xscale=10]
    \draw[thick] (0,0) -- (0.1,0); 
    \filldraw[fill=white] (0.0601-0.0071,-0.1) rectangle (0.0601+0.0071,0.1);
    \draw[thick] (0.0601,-0.1) -- (0.0601,0.1);
\end{tikzpicture} & 0.0781$\pm$0.0055 \hspace{1mm}
\begin{tikzpicture}[baseline=-0.75ex,xscale=3]
    \draw[thick] (0,0) -- (0.3,0); 
    \filldraw[fill=white] (0.0781-0.0055,-0.1) rectangle (0.0781+0.0055,0.1);
    \draw[thick] (0.0781,-0.1) -- (0.0781,0.1);
\end{tikzpicture} \\ 
RNN & 511 & 21K & 0.0270$\pm$0.0022 \hspace{1mm}
\begin{tikzpicture}[baseline=-0.75ex,xscale=10]
    \draw[thick] (0,0) -- (0.1,0); 
    \filldraw[fill=white] (0.0270-0.0022,-0.1) rectangle (0.0270+0.0022,0.1);
    \draw[thick] (0.0270,-0.1) -- (0.0270,0.1);
\end{tikzpicture} & 0.1320$\pm$0.0127 \hspace{1mm}
\begin{tikzpicture}[baseline=-0.75ex,xscale=3]
    \draw[thick] (0,0) -- (0.3,0); 
    \filldraw[fill=white] (0.1320-0.0127,-0.1) rectangle (0.1320+0.0127,0.1);
    \draw[thick] (0.1320,-0.1) -- (0.1320,0.1);
\end{tikzpicture} \\ 
LSTM & 506 & 21K & 0.0824$\pm$0.0113 \hspace{1mm}
\begin{tikzpicture}[baseline=-0.75ex,xscale=10]
    \draw[thick] (0,0) -- (0.1,0); 
    \filldraw[fill=white] (0.0824-0.0113,-0.1) rectangle (0.0824+0.0113,0.1);
    \draw[thick] (0.0824,-0.1) -- (0.0824,0.1);
\end{tikzpicture} & 0.2196$\pm$0.0212 \hspace{1mm}
\begin{tikzpicture}[baseline=-0.75ex,xscale=3]
    \draw[thick] (0,0) -- (0.3,0); 
    \filldraw[fill=white] (0.2196-0.0212,-0.1) rectangle (0.2196+0.0212,0.1);
    \draw[thick] (0.2196,-0.1) -- (0.2196,0.1);
\end{tikzpicture} \\
\bottomrule
\end{tabular}

\begin{tabular}{p{10cm}p{3.8cm}p{3.8cm}}
& \begin{tikzpicture}[baseline=-0.75ex,xscale=10]
    \draw[thick] (0,0) -- (0.1,0);
    \foreach \x/\xtext in {0, 0.1} {
        \draw (\x,0.05) -- (\x,-0.05);
    }
    \node[below] at (0,-0.1) {0};
    \node[below] at (0.1,-0.1) {0.1};
\end{tikzpicture} & \begin{tikzpicture}[baseline=-0.75ex,xscale=3]
    \draw[thick] (0,0) -- (0.3,0);
    \foreach \x/\xtext in {0, 0.3} {
        \draw (\x,0.05) -- (\x,-0.05);
    }
    \node[below] at (0,-0.1) {0};
    \node[below] at (0.3,-0.1) {0.3};
\end{tikzpicture} \\
\end{tabular}
\end{table*}
\begin{figure*}[!htbp]
    \centering
    \begin{subfigure}[b]{0.23\textwidth}
        \centering
        \includegraphics[width=\textwidth, height=0.122\textheight]{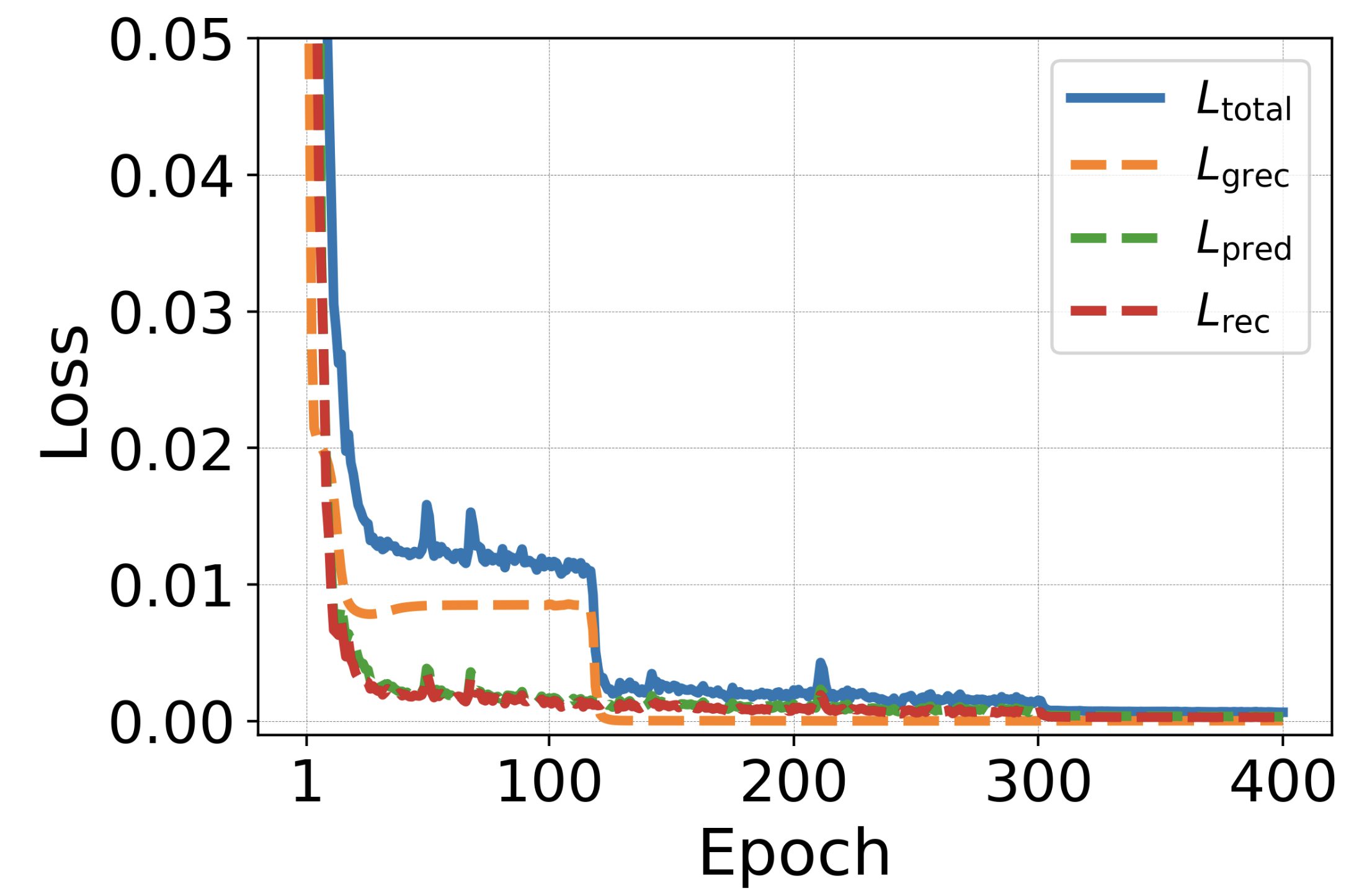}
        \caption{Loss curve evolution during training, including individual loss components.}
        \label{fig:sub1}
    \end{subfigure}
    \hspace{0.5em} 
    \begin{subfigure}[b]{0.23\textwidth}
        \centering
        \includegraphics[width=\textwidth, height=0.12\textheight]{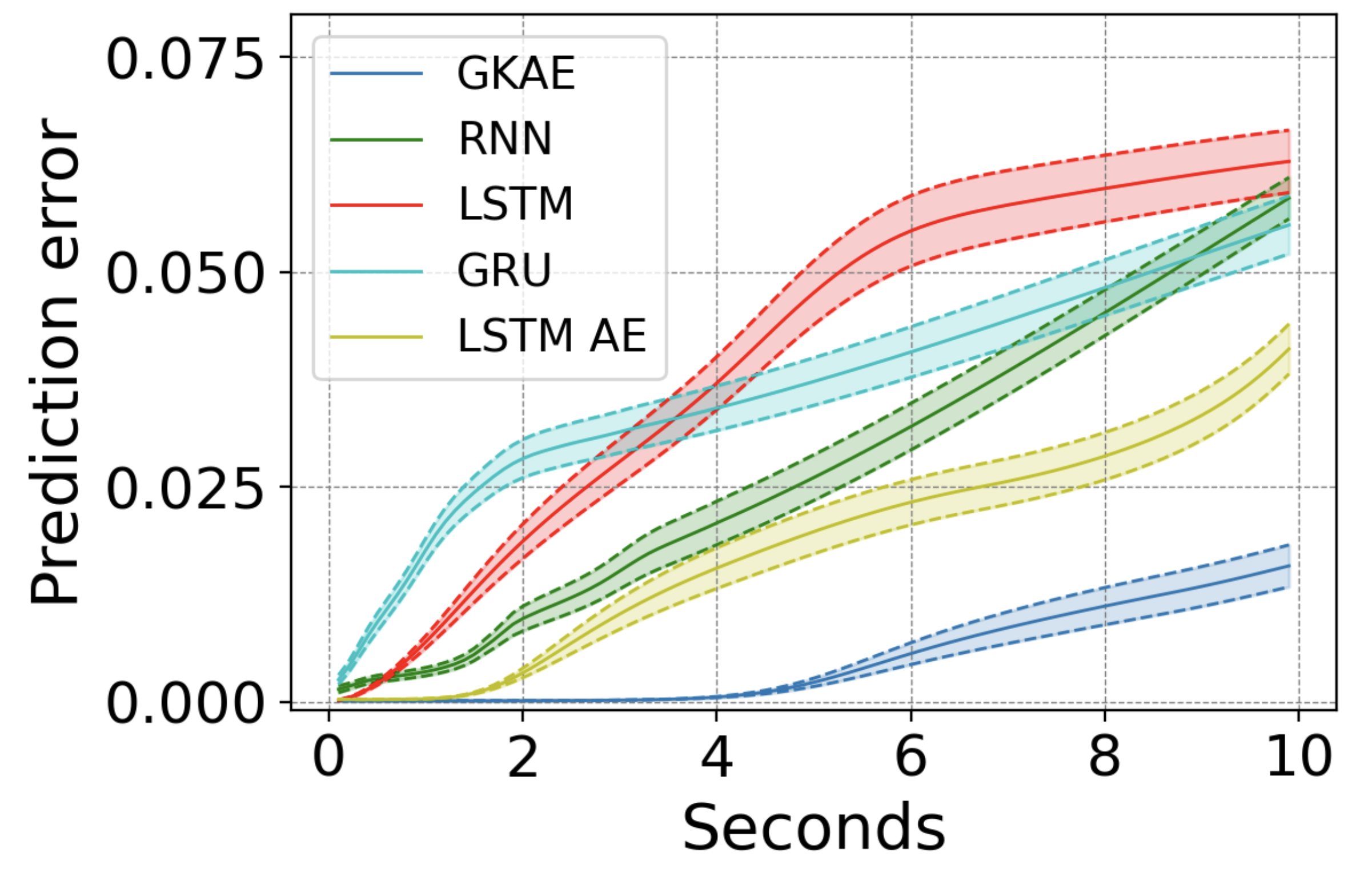}
        \caption{Evolution of prediction error over time compared to baseline methods.}
        \label{fig:sub2}
    \end{subfigure}
    \hspace{0.5em} 
    \begin{subfigure}[b]{0.23\textwidth}
        \centering
        \includegraphics[width=\textwidth, height=0.122\textheight]{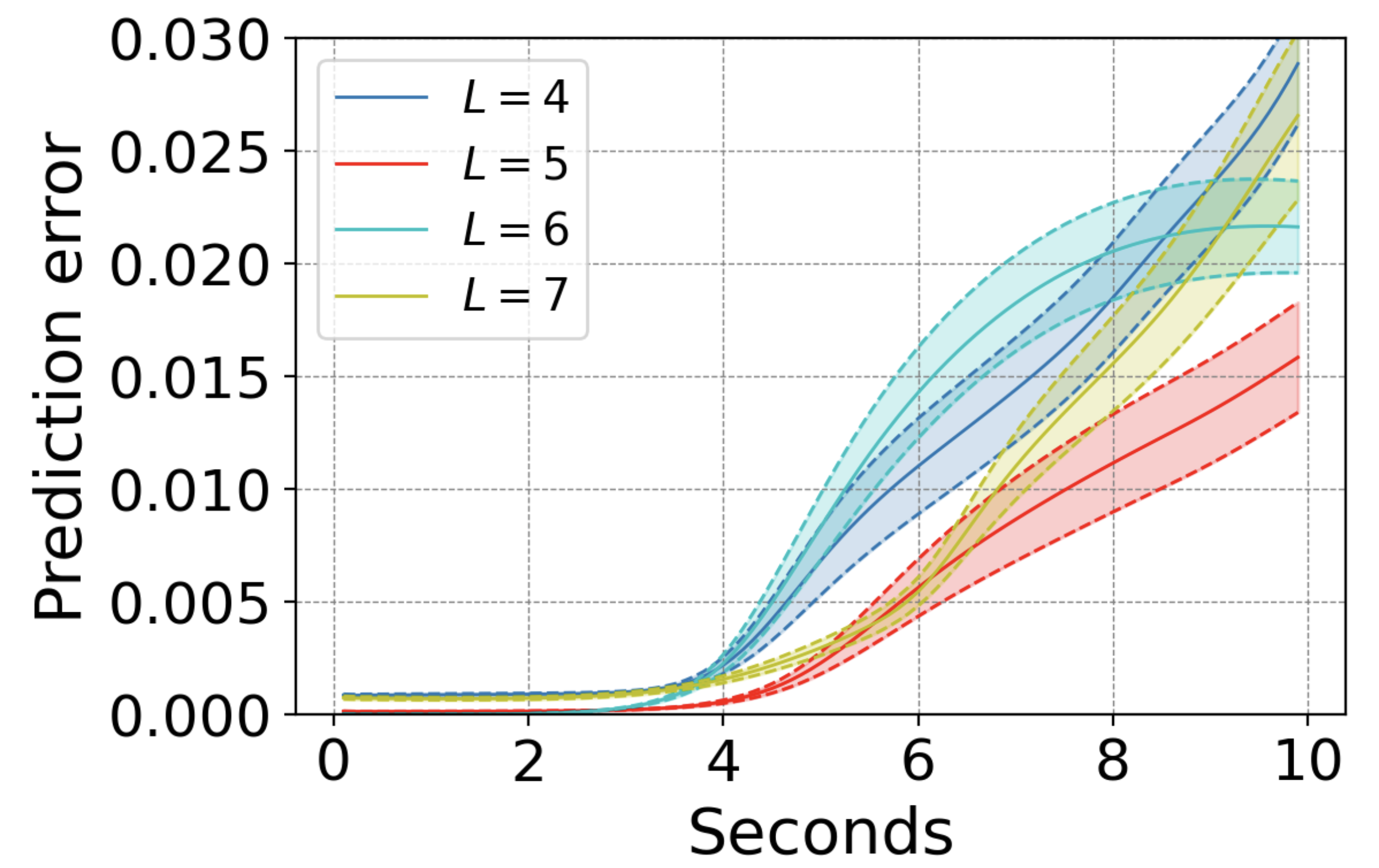}
        \caption{Prediction error evolution with varying numbers of UAVs ($L$), showing error changes.}
        \label{fig:sub3}
    \end{subfigure}
    \hspace{0.5em} 
    \begin{subfigure}[b]{0.23\textwidth}
        \centering
        \includegraphics[width=\textwidth, height=0.12\textheight]{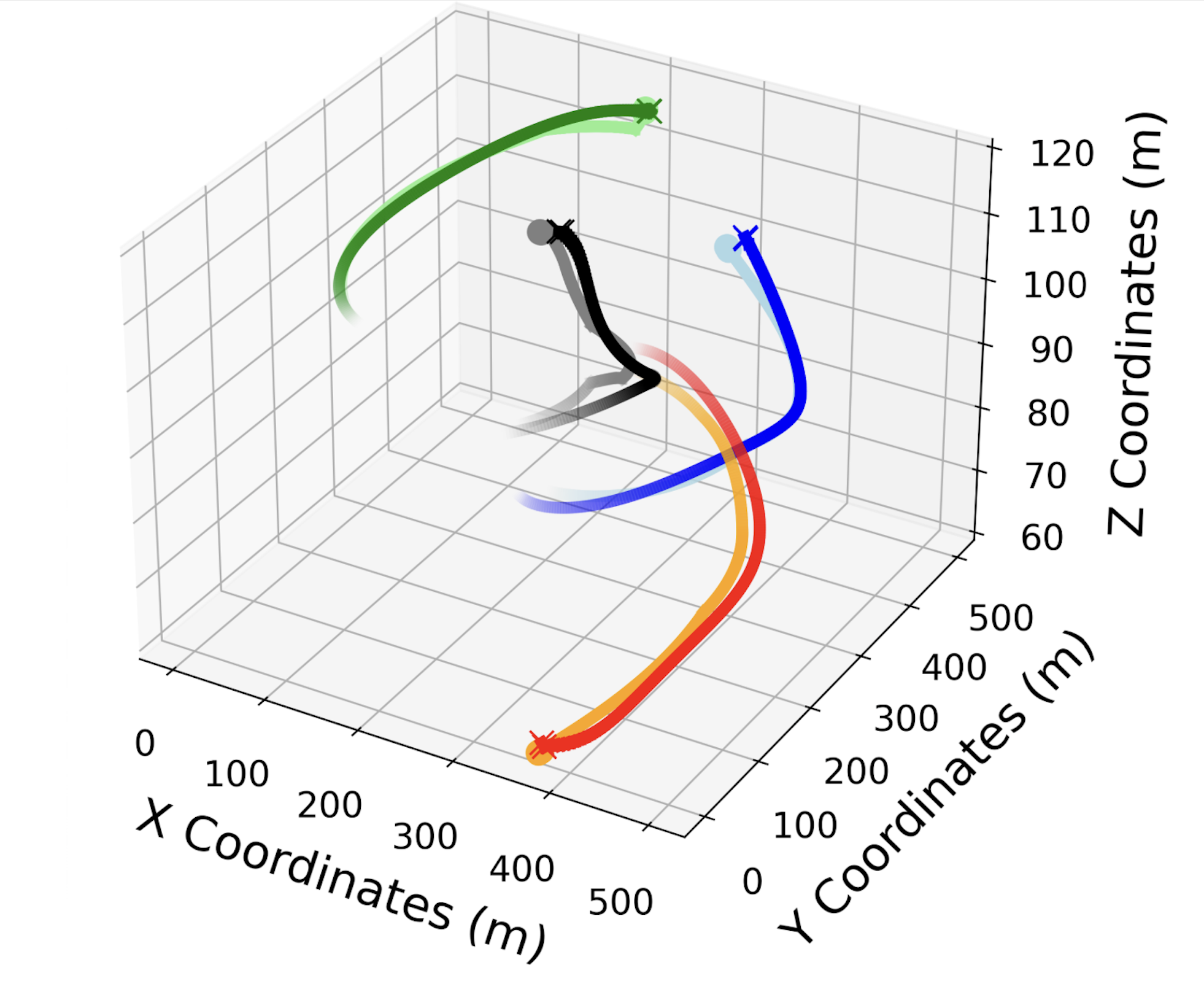}
        \caption{3D visualization of predicted trajectories. Ground-truth is lighter, predictions are darker.}
        \label{fig:sub4}
    \end{subfigure}
    \caption{Performance evaluation of the proposed model, including training loss curves and prediction error comparisons.}
    \label{fig:main}
\end{figure*}

We define the \emph{probability of detection} by the UAVs as a metric for evaluating our the success of our predictive covert operations. Then, detection event for a node $n$, can be expressed as:
\be 
\cY_{\text{det}, n}(t+\Delta t) = \{P_n(t+\Delta t) < \lambda \hat{P}_n(t+\Delta t)\}.
\ee
Considering there are $N$ ground nodes, the occurrence of a detection event in any node will signal potential suspicious activities to the UAV:
\be
\cY_{\text{det}}(t+\Delta t) = \bigcup_{n = 1}^N \cY_{\text{det}, n}(t+\Delta t).
\ee
Therefore, the probability that the event is detected at a specific time $\Delta t \in \{1, \cdots, H\}$ during the covert operation is 
\be 
\Pr (\cY_{\text{det}}(t + \Delta t)) = 
\Pr \left(\bigcup_{n = 1}^N \{P_n(t + \Delta t) < \lambda \hat{P}_n (t + \Delta t)\}\right).
\ee

Finally, the probability of detection over the prediction time window $P$ is given by
\be 
\mathds{P}_{\rm det} (N, P, \lambda)
= 
\Pr \left( \bigcup_{\Delta t = 1}^H \cY_{\text{det}}(t + \Delta t) \right).
\ee

\section{Results}\label{sec:res}
\subsection{Predicting Multi-UAV Trajectories}
In this section, we present the performance the proposed method for predicting the UAV dynamics. Using Algorithm~\ref{Algorithm:dynamics}, we create 2-dimensional dynamics and 3-dimensional dynamics. Unless specified otherwise, the simulation results report the performance when using the 3-dimensional dynamics. 
\subsubsection{Training Convergence}
In Figure~\ref{fig:sub1}, we illustrate the training loss curve of our proposed method, GKAE, showcasing both the total weighted loss and its individual components. Notably, we train our proposed method for 400 epochs and, as seen, the method converges around the $300-$th epoch. This rapid convergence can likely be attributed to the model learning the loss with a relatively small number of defined parameters, specifically 2K. 
\subsubsection{Prediction Accuracy}
In Table.~\ref{tab:training_parameters}, we compare the performance of our proposed method with the aforementioned baseline methods over 2-dimensional and 3-dimensional input, respectively. We report the prediction error aggregated over a 10 second horizon, using different initial locations for each run. For the 2D dynamics, our proposed method outperforms other methods significantly, with a $90.2\%$ decrease in mean prediction error compared to the second-best method, LSTM-AE, which has a mean error of $0.0112$. Specifically, GKAE achieves a mean error of $0.0011$. For the 3D dynamics, our proposed method demonstrates a $84.6\%$ decrease in mean prediction error, achieving a mean error of $0.005$, while the next best method, LSTM-AE, shows a mean error of $0.0324$. Notably, GKAE achieves the lowest minimum prediction error in both 2D and 3D cases, indicating its robustness and accuracy. Additionally, our method requires significantly fewer parameters and training epochs compared to the other baseline approaches. 
In Fig.~\ref{fig:sub2}, we present a comparison of the prediction error evolution and its 95\% confidence interval for our proposed methods versus baseline methods for 10 second prediction horizon, aggregated from 400 runs. It is evident that as the value of $\Delta t$ increases, the prediction error also increases. However, our proposed method consistently achieves superior performance in both short-term and long-term predictions compared to the alternatives.
In Fig.~\ref{fig:sub3}, we illustrate the prediction error of the proposed method, GKAE, and demonstrate its scalability. The 2-dimensional dynamics are defined with varying numbers of UAVs, and it can be seen that the prediction error remains comparable even as the number of UAVs increases. This illustrates that our proposed method is scalable and performs consistently well in scenarios with a high number of UAVs.

Using the trained model, we predict the next $20$ seconds of UAV trajectories, based solely on the initial location data, as seen in Fig.~\ref{fig:sub4}. It is evident that our method can accurately predict for the 3-dimensional UAV dynamics with each UAV exhibiting distinct dynamics. Using GNN, we comprehensively account for the spatial relationships inherent in Couzin dynamics. Additionally, leveraging Koopman theory aids in linearizing the non-linear trajectories, thereby resulting in long-term accurate predictions.

\subsection{Covert Communication Using Predicted Trajectories}
In this section, we assess the effectiveness of the methods in achieving a covert operation. The success of such an operation is directly linked to the accuracy of the UAV location predictions, which has been evaluated in the previous section. Our focus here is to utilize the predicted trajectories to achieve a low probability of detection. In the previous section, we establish that using the GKAE is superior to the other baselines in predicting the trajectories of the multiple UAVs. In a similar vein, we use the accurate trajectories and ensure a low probability of detection. 
\subsubsection{Compare with baselines} 
\begin{figure}[h!]
    \centering
\includegraphics[width=0.8\columnwidth]{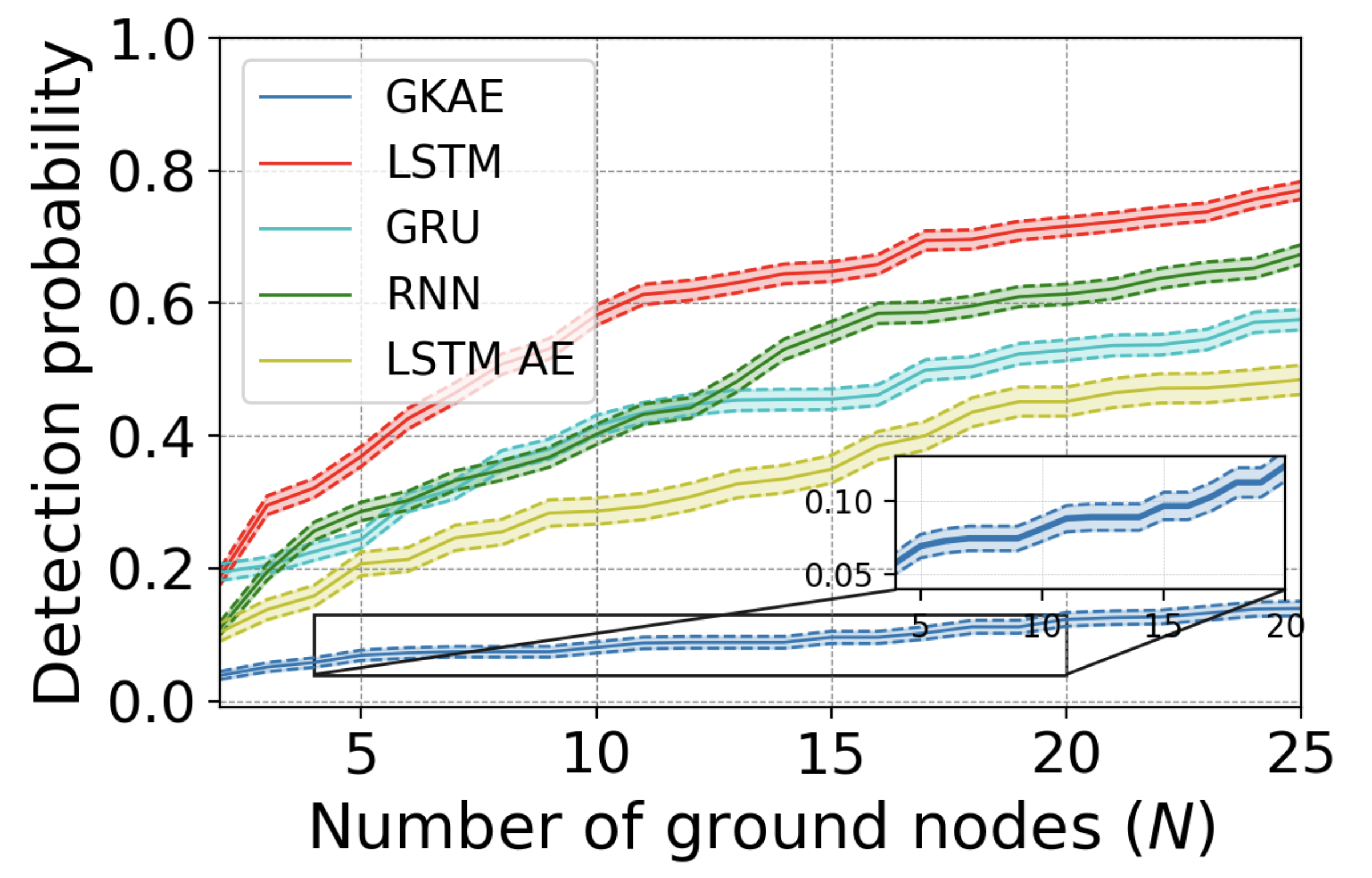}
    \caption{Comparison of the probability of detection across varying numbers of ground nodes using trajectories predicted by different methods.}
    \label{fig:vn}
\end{figure}

In Fig.~\ref{fig:vn}, we present a comparison of the probability of detection utilizing trajectories predicted by various methods discussed in the paper, over varying number of ground nodes in the covert operations. On average, it is evident that our proposed methods outperform all others in achieving a lower probability of detection when using a value of $\lambda = 0.6$. Specifically, the predicted trajectories using the GKAE method result in a 63\% to 75\% reduction in detection probability compared to the LSTM-AE, which is the second-best method. Additionally, there is a consistent increase in the probability of detection as the number of ground nodes increases due to the higher likelihood of UAV interception and detection by more ground nodes involved in the covert operation.

\subsubsection{Varying descaling factor} 
\begin{figure}[h!]
    \centering
    \begin{subfigure}[b]{0.23\textwidth}
        \centering
        \includegraphics[width=\textwidth, height=0.15\textheight]{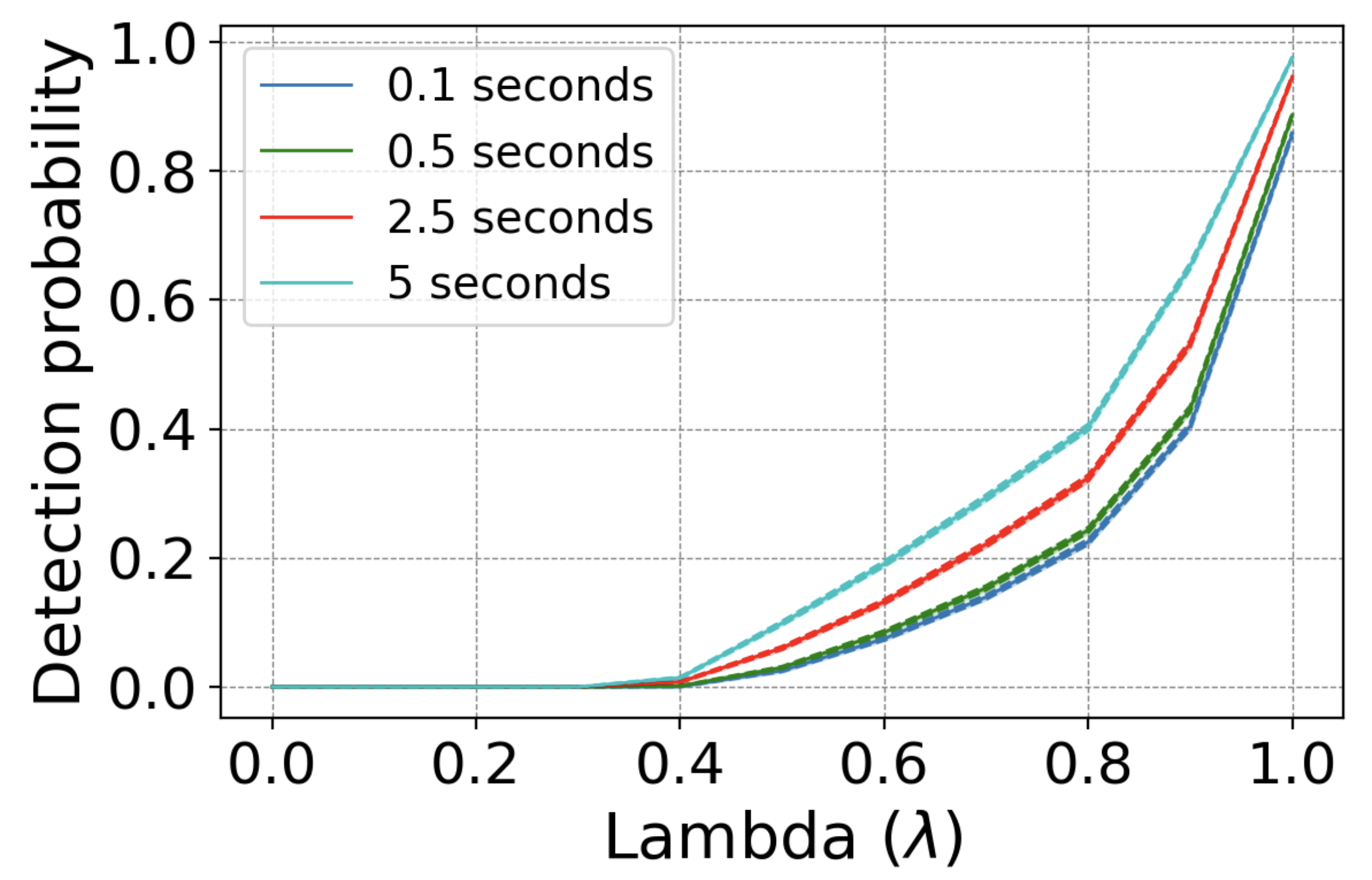}
        \caption{Comparison of the probability of detection across varying values for the descaling factor over different lengths of predictive covert operations.}
        \label{fig:lm}
    \end{subfigure}
    \hspace{0.5em} 
    \begin{subfigure}[b]{0.23\textwidth}
        \centering
        \includegraphics[width=\textwidth, height=0.15\textheight]{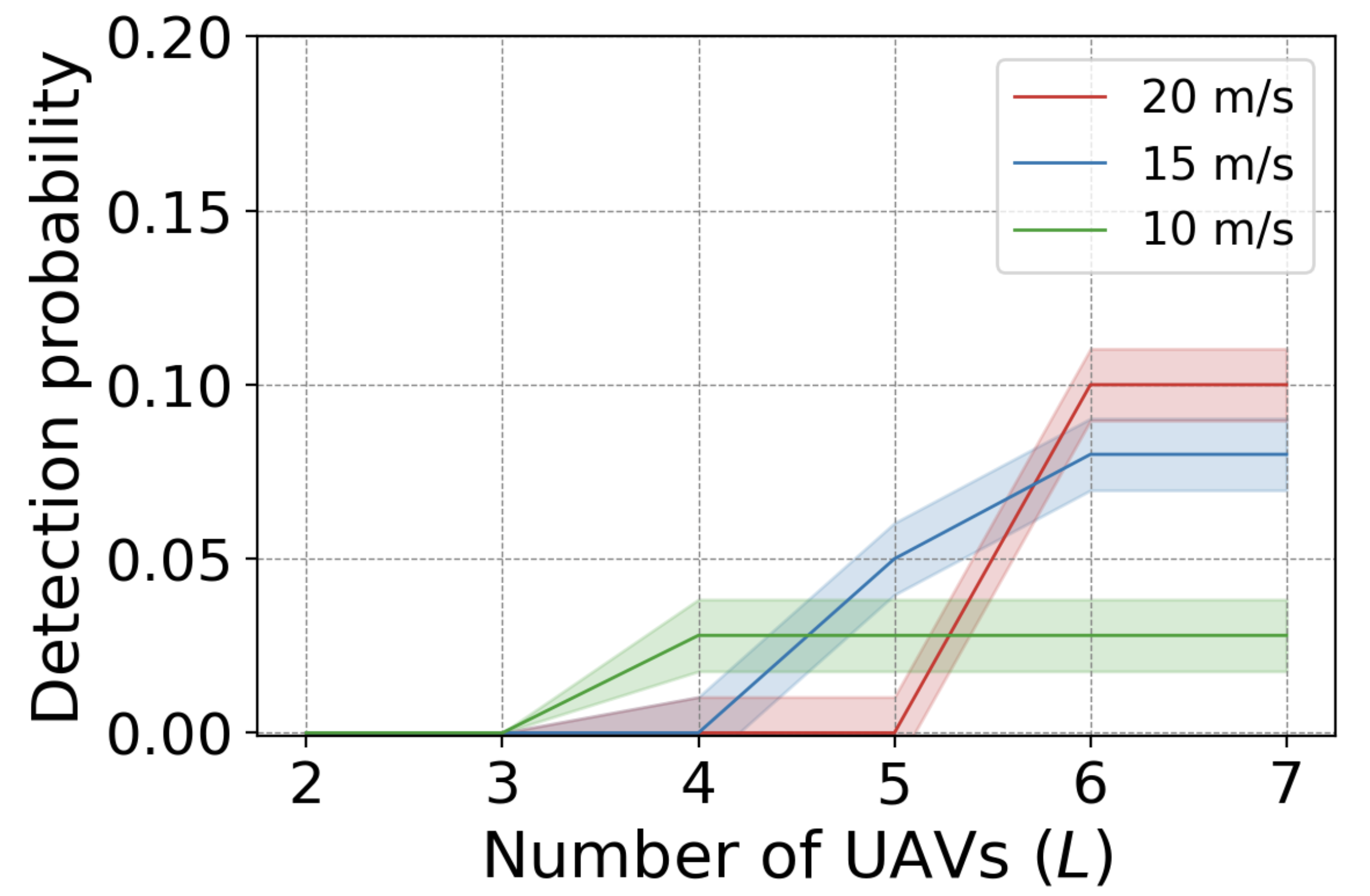}
        \caption{Comparison of the probability of detection across varying numbers of UAVs performing surveillance over different speed configurations of the UAVs.}
        \label{fig:vnu}
    \end{subfigure}
    \caption{Comparative analysis of detection probabilities: (a) descaling factor values versus operation lengths and (b) number of UAVs versus speed configurations.}
    \label{fig:combined}
\end{figure}

The success of our covert operations is dependent on the descaling factor, $\lambda$. This factor introduces a unique trade-off: higher values of $\lambda$ increase detectability while lowering the data rate, whereas smaller values reduce the probability of detection but constrain the data rate of the ground nodes. In Fig.~\ref{fig:lm},  we illustrate the average probability of detection over varying lengths of predictive covert operations. It is evident that, for all values of $\lambda$, the detection probability increases as the prediction horizon lengthens. Furthermore, as evidenced, an increase in $\lambda$ correlates with a higher probability of detection. Therefore, using a smaller value of $\lambda$ (i.e. $\lambda \le 0.5$) is preferred to achieve a balanced trade-off between detection probability and data rate, especially when the length of the predictive covert operation increases. 

\subsubsection{Varying number of UAVs} In Fig.~\ref{fig:vnu}, we analyze the effectiveness of the predictive covert operations as the number of UAVs increases over varying UAV speeds. As previously demonstrated in Fig.~\ref{fig:sub3}, our proposed method is scalable and capable of predicting the trajectories of multiple UAVs with lower prediction error. However, an increase in the number of UAVs within the operational area also increases the likelihood of detection due to the greater number of UAVs to intercept. Despite this, with accurate predictions and a descaling factor of $\lambda = 0.6$, we can maintain a low probability of detection even as the number of UAVs in the area of operation increases. On average, over varying configurations of UAV speeds, we face a 0\% to 0.0865\% chance of detection in a predictive covert operation lasting 10 seconds, the total number of UAVs is 2 and 7, respectively. 





\section{Conclusions}\label{sec:Conc}

In this paper, we studied LPD communication for terrestrial ad-hoc networks under UAV surveillance using a data-driven technique to predict the locations of UAVs. We considered a specific scenario where a central unit monitors a group of surveillance UAVs from a remote location and provides crucial information to the nodes of ad-hoc networks for covert communication by learning and predicting the trajectory of UAVs. To fully understand the UAV mobility pattern, we aimed to predict the trajectory of UAV surveillance using a novel data-driven approach that integrates graph learning with Koopman theory. By leveraging the GNN architecture, we were able to learn the intricate spatial interactions within the UAV network and linearize the dynamics of multiple UAVs using the same architecture. Utilizing these predicted locations, we conducted a case study to optimize the nodes' transmit power in a terrestrial ad-hoc network, minimizing the detectability of RF signals. Extensive simulations demonstrated accurate long-term predictions of UAV trajectories, which were also compared with those of some well-known baseline techniques, showing promise for enabling low-latency LPD covert operations.

\bibliographystyle{ieeetr}
\bibliography{Koopman}

\end{document}